\documentclass[iicol,sn-mathphys-num]{sn-jnl}
\pdfoutput=1

\usepackage{graphicx}
\usepackage{multirow}
\usepackage{amsmath,amssymb,amsfonts}
\usepackage{amsthm}
\usepackage{mathrsfs}
\usepackage[title]{appendix}
\usepackage{xcolor}
\usepackage{textcomp}
\usepackage{manyfoot}
\usepackage{booktabs}

\newcommand{\insetfont}{\small}

\begin{document}

\title{FROG: A new people detection dataset for knee-high 2D range finders}

\author*[1]{\fnm{Fernando} \sur{Amodeo}}\email{famozur@upo.es}
\equalcont{These authors contributed equally to this work.}
\author[1]{\fnm{No\'e} \sur{P\'erez-Higueras}}\email{noeperez@upo.es}
\equalcont{These authors contributed equally to this work.}
\author[1]{\fnm{Luis} \sur{Merino}}\email{lmercab@upo.es}
\author[1]{\fnm{Fernando} \sur{Caballero}}\email{fcaballero@upo.es}

\affil[1]{\orgdiv{Service Robotics Laboratory}, \orgname{Universidad Pablo de Olavide}, \orgaddress{\city{Seville}, \country{Spain}}}

\abstract{

    Mobile robots require knowledge of the environment, especially of humans located in its vicinity. While the most common approaches for detecting humans involve computer vision, an often overlooked hardware feature of robots for people detection are their 2D range finders. These were originally intended for obstacle avoidance and mapping/SLAM tasks. In most robots, they are conveniently located at a height approximately between the ankle and the knee, so they can be used for detecting people too, and with a larger field of view and depth resolution compared to cameras.

    In this paper, we present a new dataset for people detection using knee-high 2D range finders called FROG.
    This dataset has greater laser resolution, scanning frequency, and more complete annotation data compared to existing datasets such as DROW \cite{Beyer2018RAL}.
    Particularly, the FROG dataset contains annotations for 100\% of its laser scans (unlike DROW which only annotates 5\%), 17x more annotated scans, 100x more people annotations, and over twice the distance traveled by the robot.
    We propose a benchmark based on the FROG dataset, and analyze a collection of state-of-the-art people detectors based on 2D range finder data.
    
    We also propose and evaluate a new end-to-end deep learning approach for people detection. Our solution works with the raw sensor data directly (not needing hand-crafted input data features), thus avoiding CPU preprocessing and releasing the developer of understanding specific domain heuristics. Experimental results show how the proposed people detector attains results comparable to the state of the art, while an optimized implementation for ROS can operate at more than 500 Hz.

}


\maketitle

\section*{Acknowledgements}

\textbf{Funding:} This work is partially supported by the project PICRAH4.0 (PLEC2023-010353) funded by programa Transmisiones 2023 del Ministerio de Ciencia e Innovaci\'on, and by the project NORDIC (TED2021-132476B-I00) funded by MCIN/AEI/10.13039/501100011033 and the European Union ``NextGenerationEU''/``PRTR''.

\noindent \textbf{Data availability:} The FROG dataset is available at: \url{https://robotics.upo.es/datasets/frog/laser2d_people/}.
The dataset is based upon prior published work \cite{Ramon2014}, and no personally identifying data is present due to the nature of 2D LiDAR data.

\noindent \textbf{Ethics declaration:} The authors have no conflicts of interest regarding this article. 

\section{Introduction}

Nowadays, mobile robots are becoming part of our daily lives. Robots must be capable of sharing the space with humans in their operational environments. Therefore, human social conventions must be taken into account when navigating within the scenario in order to improve people's comfort. The first step to achieve this is human perception. Robots must be able to detect people in their surroundings, distinguishing them from other static
and dynamic obstacles.

In the last few years, image-based algorithms for detection and tracking of people have evolved significantly. Moreover, these algorithms can also work in 3D space by using cameras with depth perception. However, the use of these cameras for human detection in the robot navigation task still presents some drawbacks. The field of view of most cameras is very limited, and so is the depth perception range. Robots usually work around these limitations by making use of several cameras, which thus increases the complexity of the system and the computation requirements.

On the other hand, most commercial and non-commercial ground robots include 2D LiDAR range finders.
This includes industrial robots used in warehouses such as autonomous mobile robots (AMRs) and automated guided vehicles (AGVs), which must perform people detection if their specification calls for sharing the operating environment with human workers.
In any case, another major driving motivation for 2D LiDAR usage is reducing the total cost of the robot, while still providing a platform that can reliably perform
obstacle detection and robot localization.
The cost of 2D LiDAR sensors has gone down in recent years, and
economical models for hobbyist/educational use can even be found in mainstream
marketplaces.
In addition,
LiDARs provide accurate range measurements closely achieving full $360^\circ$ coverage as well as long range detection, making them a good choice for the aforementioned use cases.
Furthermore, in most cases, these sensors are installed on robots at a plane parallel to the ground that is approximately knee height.

A number of robotics researchers have worked on using 2D range finders to detect people in the proximity of a robot. The first approaches used hand-crafted features and classical algorithms \cite{Arras2007UsingBF,RosLeg2010R}, while later approaches employed deep learning techniques \cite{BeyerHermans2016RAL,Beyer2018RAL,petra2019,Jia2020DRSPAAM,Jia2021Person2DRange}. However, most publicly available datasets for people detection in robotics involve other kinds of sensors, or require relabeling. Very few datasets specifically geared for 2D range finders exist, the most notable of which is the DROW dataset \cite{BeyerHermans2016RAL,Beyer2018RAL}.

This work aims to fill this gap by releasing a completely new 2D range finder dataset specifically focused on person detection.
The laser scans were recorded as part of the Fun Robot Outdoor Guide (FROG) project \cite{hri2014frog}.
The scenario consists of a tour of the Royal Alc\'azar of Seville,
an iconic Mud\'ejar palace receiving over 1.5 million visitors a year.
Our dataset contains a large number of laser scans, and unlike \cite{BeyerHermans2016RAL} every single one is annotated.
This is possible using our semi-automatic annotation tool, which considerably reduces the workload required to annotate such an extensive dataset.

Overall, we present the following contributions:

\begin{itemize}
    \item A fully annotated 2D range finder person detection dataset including
    a variety of indoors and outdoors scenes, crowded scenes, and challenging
    features (such as pillars, bushes, slopes, etc).
    \item A deep learning based people detection model that learns
    people-distinguishing features directly from the range data vector without
    requiring a preprocessing step,
    and produces person location proposals using techniques analogous to
    those of image-based object detectors.
    Our ROS-based implementation also achieves faster inference times than other currently available solutions.
    \item A benchmarking codebase and methodology based around our dataset,
    which allows researchers to evaluate their own 2D range finder person
    detectors under standardized metrics and metric computation code.
\end{itemize}

\section{Related work}

In this section we survey existing datasets related to people detection and range finders, with attention to their composition and attributes.
We also survey existing works that aim to detect people using these sensors. Our findings, detailed below, show that most datasets are either geared towards autonomous driving tasks (with limited genericity and relevance to people detection), or involve other kinds of sensors. People detectors also tend to be based on classical algorithms or make use of hand-crafted input processing. We focused on studying detectors involving the use of deep learning, even if they also contain non-deep processes.

\subsection{Annotated 2D LiDAR datasets}

There are many datasets with people annotations in the form of bounding boxes or segmentation masks, geared towards plain 2D images or sensors such as RGBD cameras or 3D LiDAR.
However, there is a scarcity of people detection datasets geared towards 2D LiDAR sensors, containing annotated 2D range data.

With the rise of autonomous (self-driving) cars, several multimodal datasets have been recorded and released, such as nuScenes \cite{nuscenes2020}, KITTI \cite{kitti2013}, and PedX \cite{kim2019pedx}. These datasets focus on traffic scenes, where most objects on the roads are vehicles, and pedestrians are sparsely distributed.

Focused on common pedestrian situations in indoor and outdoor environments, we found datasets with pedestrians annotated in images and 3D LiDAR point clouds like JRDB \cite{MartnMartn2021JRDBAD}, SCAND \cite{SCANDKarnan22}, STCrowd \cite{CongSTCrowd2022} or WILDTRACK \cite{8578626} (the latter only based on static camera images in outdoor scenarios). There also exist 2D LiDAR datasets for general purpose segmentation such as Semantic2D \cite{semantic2d}, however they contain many more classes besides ``person'', and thus they are not well suited for training people detectors.

Other datasets provide annotated trajectories of human pedestrians performed in a unique controlled laboratory environment like the THÖR dataset \cite{thorDataset2020} or the Magni Human Motion dataset \cite{SchreiterMagniRoman2023}. These datasets are dedicated to learning social navigation (as opposed to simply people detection), and only the latter provides data from a robot's mounted 2D LiDAR sensor.

In this work, we focus on datasets with annotated pedestrians in 2D laser scans, and people detectors based on the same sensory input.
The DROW dataset was introduced in \cite{BeyerHermans2016RAL} for the detection of wheelchairs (\texttt{wc}) and walking aids (\texttt{wa}) in laser scan data. The authors recorded 113 sequences at an elderly care facility. In a follow-up work \cite{Beyer2018RAL}, the authors added person (\texttt{wp}) annotations. We found the following drawbacks in the dataset:

\begin{itemize}
    \item During the annotation process, the scans were batched in groups of 100, and only 1 out of 4 batches was provided to human volunteers. Moreover, within each batch, only 1 out of 5 scans was annotated. This combination results in just 1 out of 20 scans (5\% of the total) carrying annotations, with the remaining 95\% being left completely unannotated.
    Even though the authors justified
    this decision in reducing the workload of the annotators, as well as reserving
    the unannotated scans within each batch for temporal approaches;
    it still means a large majority of the data is unusable for direct supervised learning, reducing the variability of input samples and prompting the use of data augmentation.
    In addition, temporal approaches such as DR-SPAAM \cite{Jia2020DRSPAAM} do not necessarily follow
    the prescribed temporal window stride hyperparameter, instead experimenting with
    different strides (such as $T = 10$).
    \item Despite the authors' efforts in adding people annotations, the dataset
    is still mainly focused on detecting mobility aids, meaning the amount and
    quality of person annotations is inadequate for other use cases, compromising
    the genericity of the dataset.
    \item As pointed out by \cite{Jia2020DRSPAAM}, the validation set is considerably
    more challenging than the training or test sets because it contains more people
    annotations at farther distances (meaning sparser points). This causes problems
    during hyperparameter search, and can also lead researchers to make mistakes when trying to assess any possible overfitting.
\end{itemize}

Another recently available 2D LiDAR dataset for people detection is Sixth Sense \cite{sixthsense}. This dataset leverages an additional Azure Kinect sensor to produce unsupervised people detections. However, the dataset is very short (around 55k scans at 10 Hz), and it is recorded fully indoors within a university campus. Moreover, the person detection data is not directly in the form of annotations, instead being presented as the output of 360$^{\circ}$ human distance/presence detectors; thus requiring further processing to separate each person instance and generate person annotations usable with existing detectors.

\subsection{People detectors}

There are plenty of different detectors and trackers of people based on images
and depth perception, as commented previously. These include face detectors,
full-body detectors, or even skeleton detectors.
However, the field of people detection in 2D range data has not been thoroughly
explored and researched.
We consider this to be related to the complexity of the problem, given the scarcity
of reliable information that can be extracted from the range data in order to
detect people.

Arras et al. developed a segment-based classifier that detected people's legs using hand-crafted features extracted from each segment \cite{Arras2007UsingBF}.
Later, an implementation of Arras's leg detector classifier was released for its use with
the ROS middleware \cite{RosLeg2010R}.

Particularly, we are interested in more novel approaches based on Deep Learning.
The PeTra (People Tracking) detector \cite{petra2019} replaced the shallow learning based algorithm
in Arras's leg detector with a deep 2D fully convolutional segmentation network
(using a projected 2D occupancy map of the range data as input),
while still maintaining a classical post-processing step for extracting locations
of individual legs from the segmentation output, as well as combining legs into person
detections. The authors also propose using a Kalman filter to produce smoother
person tracking over time.

The DROW (Distance RObust Wheelchair/Walker) detector \cite{BeyerHermans2016RAL,Beyer2018RAL}
proposes creating many small fixed-size windows centered around every point of the
scan called ``cutouts'', which are normalized to contain the same fixed number
of range values. This eliminates the spatial density variability problem caused by
laser points at different distances. Then, a 1D convolutional network is
used to extract features from each cutout, and decide both whether a person is
nearby, as well as regress a spatial offset to said person. The regressed spatial
offsets are taken as votes, and used to refine the final detected location of the person.
The network is trained with a dataset of range data created and labeled by the same authors (DROW dataset).
The authors then followed up with an improved version of their detector that fuses
temporal information \cite{Beyer2018RAL}, and spatially aligning cutouts with those
of recent past scans with the help of odometry data from the robot.

A newer people detector work \cite{Jia2020DRSPAAM} proposes
Distance Robust Spatial-Attention and Auto-regressive Model (DR-SPAAM).
Similar to DROW, it continues to use cutouts of the laser scan data,
but also using a forward looking paradigm to aggregate temporal information.
Instead of computing spatially aligned cutouts on the past scans, it uses a
similarity-based spatial attention module, which allows the CNN to learn to
associate misaligned features from a spatial neighborhood. The same
authors also present a self-supervised approach of the DR-SPAAM detector
\cite{Jia2021Person2DRange} in which a calibrated camera
with a conventional image-based object detector model
is initially used to detect the people in the scene, and subsequently used to
generate ``pseudo-labels'' in the range data for self-supervised learning.

Finally, more recent works include Li2Former \cite{li2former}, which replaces the traditional CNN used by the cutout-based approach with a Transformer-based architecture; however at the expense of increased model and training complexity, and heavier runtime processing leading to decreased speed. Moreover, to the best of our knowledge there are no publicly available implementations of this detector ready for use by robotics researchers.

A general trend in the surveyed detectors is the combination of a deep learning
network with non-deep pre-processing and post-processing steps. In particular,
the cutout-based approach increases the dimensionality of the input by virtue of
generating as many cutouts as there are points in each range data vector.
Likewise, the temporal approach involves aggregating data from several scans at
once, possibly with alignment steps (both deep and non-deep). These factors all
result in increased memory and computational overhead at both training and
inference time -- according to \cite{Jia2020DRSPAAM}, their baseline implementation of the DROW detector
(without temporal aggregation) needs 97.4ms per scan (10.4 FPS) on an edge device suitable for
robotics (Jetson AGX), while a special ``faster'' implementation of DR-SPAAM called
``DR-SPAAM*'' needs 44.3ms per scan (22.6 FPS). One of the motivations of this work is encouraging
further research into removing the need for these non-deep steps, and specifically
in Section~\ref{sect:ourowndet} we will propose an initial approach into a fully
deep person detector based on 2D range data.

\section{FROG dataset} \label{sec:frogdataset}

The FROG dataset is a large dataset of people detection in 2D LiDAR data covering a populated public space, the Royal Alc\'azar of Seville. The Alc\'azar is a UNESCO World Heritage Site famous for its Mud\'ejar Hispano-Muslim
architecture and its verdant gardens and courtyards, which receives over 1.5 million visitors a year -- one of the most visited monuments in Spain. As a result, the dataset presents a rich variety of highly populated areas and scenarios, both indoors and outdoors.

The FROG dataset for 2D laser people detection is available for download from our
website\footnote{\url{https://robotics.upo.es/datasets/frog/laser2d_people/}}.
The source of the data is our previous dataset \cite{Ramon2014}, which contains
a larger collection of raw sensor data
appropriate for localization and human-robot interaction purposes.
This data was also previously made publicly available\footnote{\url{https://robotics.upo.es/datasets/frog/upo/}}.

\begin{figure}[t]
    \centering
    \insetfont
    \includegraphics[width=0.3\textwidth]{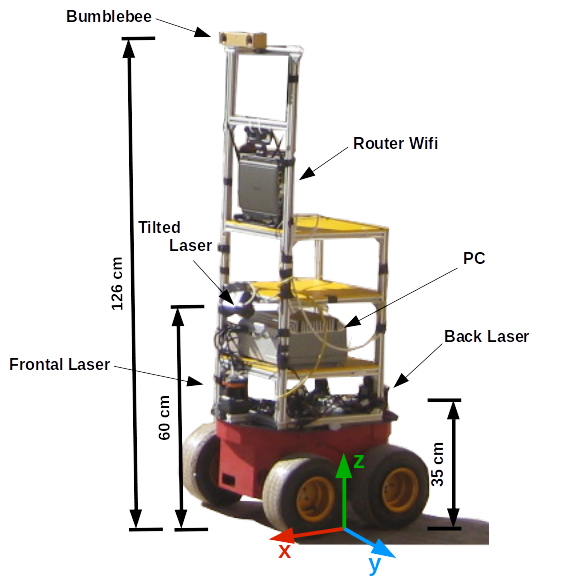}%
    \includegraphics[width=0.18\textwidth]{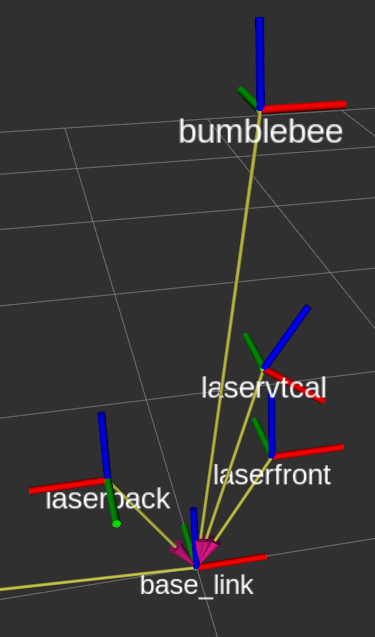}
    \caption{
    Left: image of the robot platform used for recording.
    Right: reference frames of the robot. The front mounted 2D LiDAR sensor (\texttt{laserfront}) is placed at X = 0.22\,m and Z = 0.33\,m with respect to the base of the robot (\texttt{base\_link}).
    }
    \label{fig:frog_tf}
\end{figure}

The mobile robotic platform shown in Fig.~\ref{fig:frog_tf} is used to record the data of different sensors onboard. In particular, the FROG dataset provides the data of the front-mounted 2D LiDAR, along with annotations about the people in the field of view of the sensor (180$^{\circ}$), at a maximum distance of 10 m. The odometry data is also provided. 

\begin{figure}
    \centering
    \insetfont
    \includegraphics[width=0.8\columnwidth, angle=180]{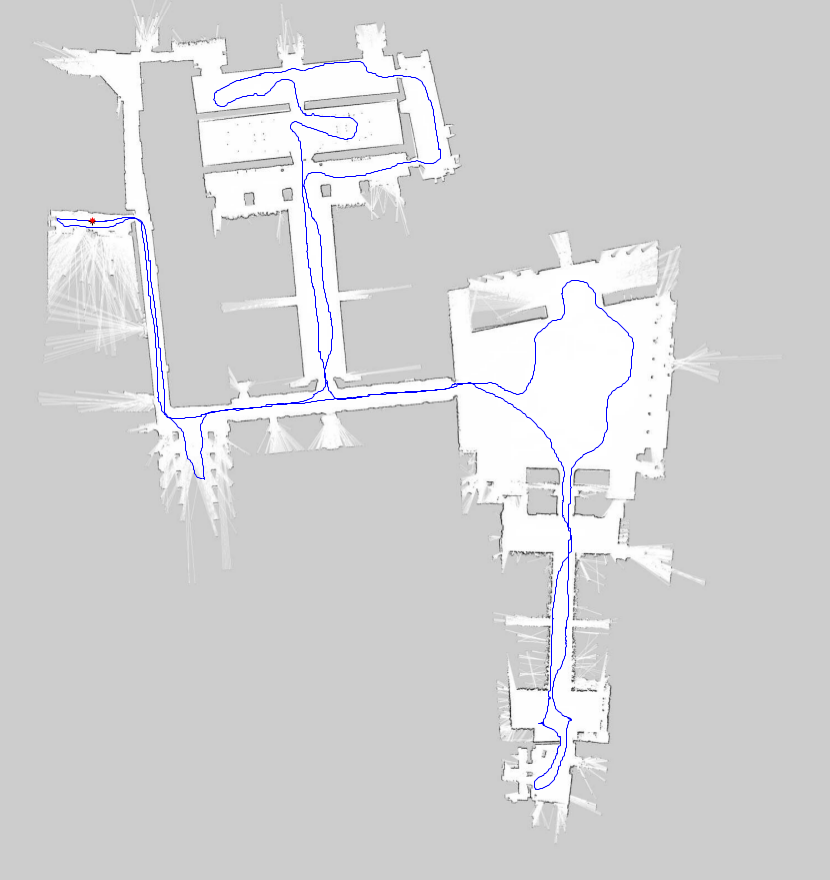}
    \caption{
        Example navigation plan used by the robot during capture of the FROG dataset.
    }
    \label{fig:frogpath}
\end{figure}

The recorded data encompasses different timetables along four days of experiments. Each sequence corresponds to a tour around the Alc\'azar. The robot trajectory of one of these tours can be seen in Fig.~\ref{fig:frogpath}. Table~\ref{tab:sessions} shows a summary of several features of the recorded sequences. 

We focus on comparing our FROG dataset against the DROW dataset, the only
currently available dataset that has been used to
evaluate and compare 2D laser based people detectors.
Table~\ref{tab:datasets} shows a detailed quantitative comparison.
Although the DROW dataset includes more hours of recordings, only a
very small portion of the data is annotated -- only 5.17\% of the scans have
associated annotations in the \texttt{.wp} files (this number includes empty lists
of people).
On the other side, the FROG dataset provides annotations for every single scan,
a richer variety of crowded scenarios, and greatly increased laser/temporal
resolutions.
Moreover, our robot was able to move at faster navigation speeds than
those achieved in DROW's scenario, and thus traversed longer paths --
over twice the distance compared to the odometry information provided
by the DROW dataset.

\begin{table*}
    \centering
    \insetfont
    \caption{
        General overview of the annotated sequences in the FROG dataset.
        Each session lasted about half an hour, and we report the total
        number of scans and annotated people in each sequence.
    }
    \begin{tabular}{r r r r r}
        \toprule
        Start time & Duration & Travel distance & Scans & People \\
        \midrule
        10:31 & 26m 42s & 1666.00 m & 64238 & 127600 \\
        11:36 & 29m 41s & 1845.66 m & 71417 & 214707 \\
        12:43 & 31m 43s & 1970.69 m & 76088 & 258298 \\
        14:57 & 29m 09s & 1824.06 m & 70062 & 133197 \\
        15:53 & 25m 16s & 1569.47 m & 60758 & 133023 \\
        16:41 & 29m 29s & 1843.57 m & 70923 & 153658 \\
        \midrule
        Total & 2h 52m 30s & 10719.45 m & 413486 & 1020483 \\
        \botrule
    \end{tabular}
    \label{tab:sessions}
\end{table*}

\begin{table*}
    \centering
    \insetfont
    \caption{
        Comparison between the DROW and FROG datasets, showing different
        general metrics about each, and also including a comparison of
        the 2D LiDAR sensor used by the robots.
    }
    \begin{tabular}{l c c}
        \toprule
        \multicolumn{1}{c}{} & \textbf{DROW} & \textbf{FROG} \textit{(ours)} \\
        \midrule
        Scenario & Elderly care facility & Royal Alc\'azar of Seville \\
        \midrule
        Total scans        & 464013        & 413486         \\
        Annotated scans    & 24012         & 413486         \\
        Populated scans & 14339 & 292889 \\
        Recorded time      & ca. 10 h      & ca. 3 h        \\
        Travel distance    & 5.18 km       & 10.72 km       \\
        People annotations & 28984         & 1020483        \\
        Avg. \# people/scan & 1.2 & 2.5 \\
        Max. \# people/scan & 17 & 16 \\
        \midrule
        Laser model        & SICK S300    & Hokuyo UTM-30LX \\
        Laser frequency    & ca. 13 Hz     & 40 Hz          \\
        Laser height       & 37 cm         & 35 cm          \\
        Laser points       & 450           & 720            \\
        Laser FoV          & 225$^{\circ}$ & 180$^{\circ}$  \\
        Laser resolution   & 0.5$^{\circ}$ & 0.25$^{\circ}$ \\
        \botrule
    \end{tabular}
    \label{tab:datasets}
\end{table*}

\subsection{Laser scan labeling tool}

\begin{figure}[t]
    \centering
    \insetfont
    \includegraphics[width=0.48\textwidth]{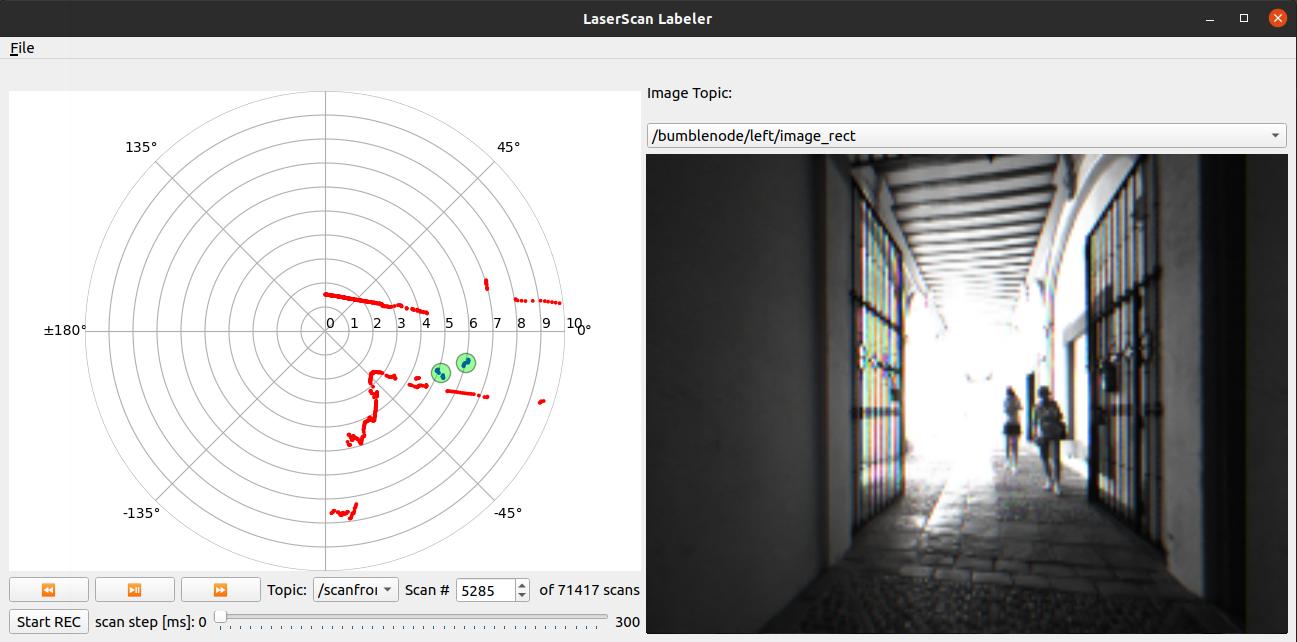}
    \caption{Main interface of the laser scan labeling tool. The tool displays the laser scan and the video feed from a camera topic side by side, and allows the user to easily create and track annotations using the mouse.}
    \label{fig:labeling_tool}
\end{figure}

Recording data from a robot typically involves using the ROS framework, which provides many facilities for interacting with device drivers, such as capturing data from the robot's sensors.
In ROS, the data of the 2D laser range finder sensors is provided through the structure given by the message \texttt{sensor\_msgs/LaserScan}. In a nutshell, the range data is expressed as an array of distances in meters. Each position in the array can be mapped to a specific laser angle using metadata included in the ROS message, specifically the minimum/maximum angles covered by the sensor and the angle increment between measures. Finally, captured data from one or more sensors is usually stored in a format known as a \textit{ROS bag file}.

We present our graphical tool used to annotate the dataset. It loads ROS bag files containing ROS laser scan messages, and graphically displays them using a top-down projection. 
The tool also allows visualizing image messages from a camera side by side (\texttt{sensor\_msgs/Image}) if they are available either in the same ROS bag file or in an external time-synchronized bag file. This can help the user identify the people to be labeled in the scene.
The scan labeler is implemented using Python 3, PyQt5 and ROS Noetic. It is publicly available on GitHub\footnote{\url{https://github.com/robotics-upo/laserscan_labeler}}, where a detailed description and instructions for use are included. 

The main interface of the application can be seen in Fig.~\ref{fig:labeling_tool}. On the left side panel we can observe the projected laser scan (in red). The scan can then be labeled by creating/removing annotation circles enclosing the laser points that correspond to each person using the mouse. The user can at any time create, move, modify the radius or delete any circle. Moreover, the tool includes several options for playing back the laser scans and moving forwards/backwards in time at different speeds. Support for additional annotation classes is also present, like baby strollers, wheelchairs or other walking aids, and intended for future use. 

An important feature of our tool is the ability to track the group of points inside each circle through time. When the user advances from one scan to the next, the center point of each circle is automatically recomputed as the mean of the points that still fall within the circle, which allows tracking each labeled person. This simple tracking does not depend on any automatic detection and is supervised by a human annotator to correct or restart it in case the tracking fails or non-person scan points are included in the circles. The tool does not enforce a minimum number of LiDAR points to create or track a circle -- the human annotator is in charge of creating or deleting them appropriately based on the LiDAR information, as well as additional information such as the visual image or intuition. Our tool also assigns an internal identifier to each tracklet that the user creates, and these identifiers are temporally consistent for each tracklet (but there are no globally consistent IDs). Thanks to this feature, the annotation circles move along with people in subsequent scans, which considerably simplifies and speeds up the annotation process. The annotators are entirely responsible for resolving edge cases. For example, when people stand close by, the annotators were instructed to shrink the circles to contain only the scan points for each individual; avoiding grouping more than one person in the same circle and handling some partial occlusions.

\begin{figure*}[ht!]
    \centering
    \insetfont
    \includegraphics[width=0.8\textwidth]{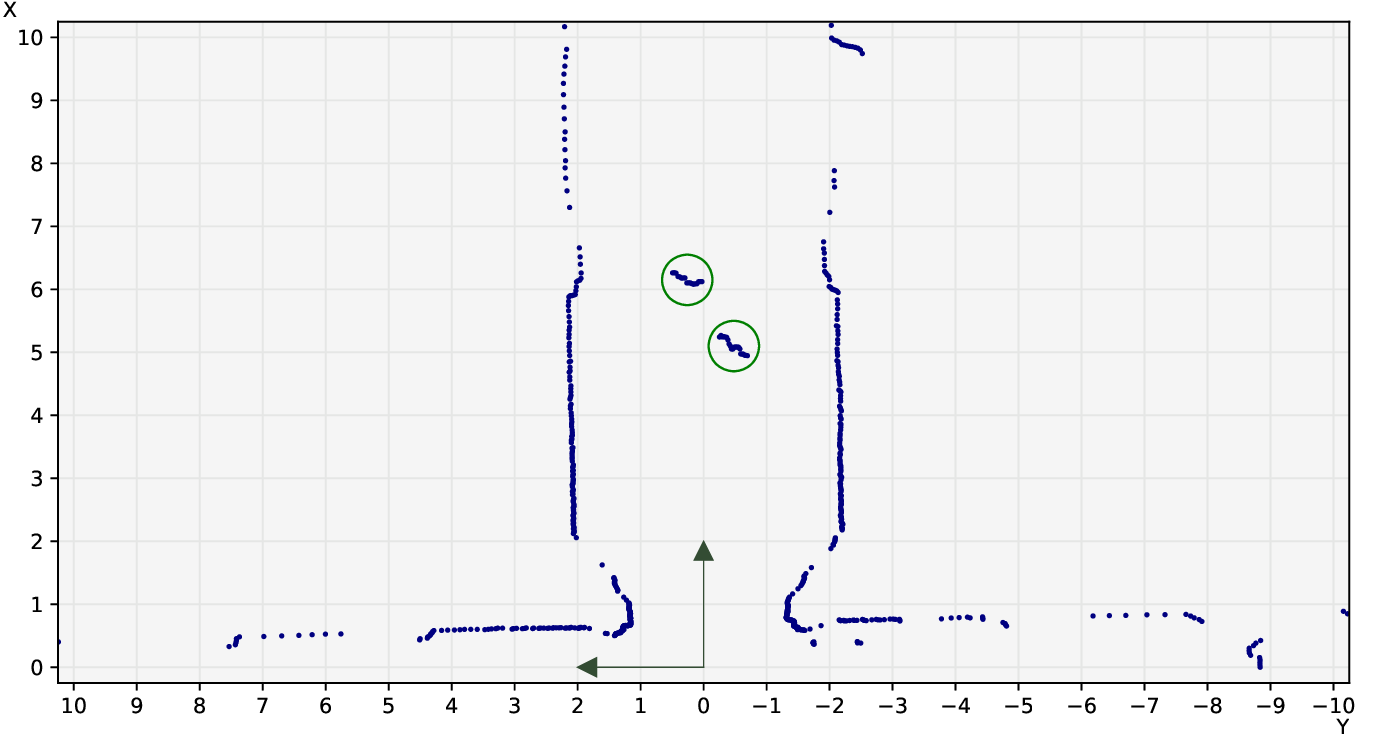}
    \caption{
        Example annotated laser scan showing the
        coordinate system used in the FROG dataset, matching the standard conventions
        used in robotics. The distances shown are in meters.
        Blue dots: points from the scan. Green circles: annotated people.
    }
    \label{fig:coords}
\end{figure*}

In the specific case of the FROG dataset, the workload is distributed across four
human annotators, all members of our laboratory group. Each annotator is in charge of annotating one or two ROS
bag files. The work is carried out using our tool, annotating each bag in several
sessions of around 10000 scans, taking breaks in between.

The output of the annotation tool is the list of circles associated with every annotated scan. Each circle includes the person tracklet identifier, center position (in Cartesian coordinates), and circle radius. Besides the list of circles, the timestamp and index of the scan within the sequence is also included so that the annotations can be traced to the original bag.
In addition, the tool also supports generating segmentation data from the circles, containing the classification of each point in the laser scan data.

Finally, the tool supports exporting the aforementioned data in different file formats:
\texttt{.csv}, \texttt{.json}, \texttt{.mat} (Matlab) and \texttt{.npy} (NumPy).
This variety of formats is intended to make human review of the data easier, as well as subsequent loading in post-processing scripts. The finalized format of the FROG dataset is explained in the following section.

\subsection{Format}

The FROG dataset is finally delivered as a series of HDF5-formatted \cite{hdf5} files.
We use this file format in order to enable greater data loading efficiency,
because HDF5 is specifically designed to store and organize large amounts of
data, supporting partial/random access and easily integrating into Python
NumPy code.

We make available a collection of Python scripts and modules in order to facilitate
the loading process, as part of our benchmarking suite described in
Section~\ref{sec:bench}. This also includes the scripts we used to process
the ROS bag files and the annotated scan data generated using our labeling
tool, as well as exporting the data into the final HDF5 files; so that other
researchers may be able to replicate our methodology with their own data.

We define the following arrays (known as \textit{datasets} in HDF5 parlance):

\begin{itemize}
    \item \texttt{scans}: This is a \texttt{float32} array of dimension
    $(N, 720)$ containing each individual laser scan vector, where $N$ is
    the total number of scans in the file. Each individual value is measured
    in meters.
    \item \texttt{timestamps}: This is a \texttt{float64} array of dimension
    $(N)$ containing the timestamp of each scan (in seconds since the UTC Unix epoch).
    \item \texttt{circles}: This is a \texttt{float32} array of dimension
    $(M, 6)$ containing all person annotations, where $M$ is the total number
    of person annotations in the file. The second dimension contains six
    values as following, specifying the position (in both cartesian and polar coordinates)
    of the person.
    \begin{itemize}
        \item 0 and 1: Specifies the X/Y position (in meters) of the person.
        \item 2: Specifies the radius (in meters) of the bounding circle that surrounds the person.
        \item 3 and 4: Specifies the angle (in radians) and distance from the origin (in meters) of the person.
        \item 5: Specifies the half-angle that covers the bounding circle when projected from the origin.
    \end{itemize}
    Person annotations are associated with only a single scan, and the exact range of entries in the \texttt{circles} array that correspond to each scan is defined by the \texttt{circle\_idx} and \texttt{circle\_num} arrays (explained below).
    \item \texttt{circle\_idx}: This is an \texttt{uint32} array of dimension $(N)$
    that specifies the index into \texttt{circles} of the first person annotation
    associated with each scan, the other annotations being stored sequentially afterwards.
    \item \texttt{circle\_num}: This is an \texttt{uint32} array of dimension $(N)$
    that specifies the total number of person annotations associated with each scan.
    \item \texttt{split}: This is an \texttt{uint8} array of dimension $(N)$
    exclusive to the file containing the benchmark training/validation sets,
    specifying which scans belong to which set (0 = training, 1 = validation).
    The suggested split roughly follows a 90:10 proportion.
\end{itemize}

An important thing to note is that we follow the standard axis convention in robotics
(see Fig.~\ref{fig:coords}).
That is, the X axis points forward, the Y axis points left, and positive angles are
counterclockwise. This causes the laser scan vectors to effectively be stored
right-to-left.

Besides the HDF5-formatted files, we also make available the raw CSV files created with our labeling tool. These files contain partial people tracklets, meaning a single person may be associated with multiple tracklets depending on occlusions and other factors that affect the labeling process. Although existing 2D LiDAR people detection works (including this work) do not make use of them, we believe they may be useful to future researchers interested in the tracking approach.

Finally, we also make available the odometry data from each session as
separate files in \texttt{.npz} (compressed NumPy) format. Each file contains two
arrays, \texttt{ts} (containing timestamps) and \texttt{data} (containing X-position,
Y-position and Z-rotation odometry samples for each timestamp).
Like \cite{BeyerHermans2016RAL}, the values are relative to an arbitrary initial state
of the robot -- only the differences between samples are meaningful.
The odometry samples are not aligned with the scans due to differences in sample rate,
and it is up to downstream users to devise a way to interpolate the state of the robot
at each scan timestamp. Users should also keep in mind the relationship between
the base frame of the robot, and the mounted laser frame, as explained in
Section~\ref{sec:frogdataset}.

\section{People detection} \label{sect:ourowndet}

In addition to the FROG dataset, we propose a new end-to-end
deep learning network that can detect people from 2D laser scan data.
This network is inspired by image-based object detection networks such as
Faster-RCNN \cite{fasterrcnn2015} or the YOLO \cite{yolo2016} family of detectors,
and motivated by the lack of approaches that are fully based on deep learning,
instead relying on hand-crafted (non-deep) pre-processing and post-processing steps that need to
be performed outside accelerators (GPU and TPU), such as the cutout generation and vote aggregation processes
introduced by \cite{BeyerHermans2016RAL}.
We theorize those non-deep processes to be a source of processing speed bottlenecks
(especially when performed on weak edge CPUs), which limits their ability to
run on a robot's built-in hardware.
There are two contributions in our proposal: a network that can learn to extract
features from 2D range finder data for use with downstream tasks, and a grid-based
people detection head similar to RPN \cite{fasterrcnn2015}.

\subsection{Laser Feature Extractor (LFE)}

\begin{figure*}[t]
    \centering
    \insetfont
    \includegraphics[width=0.98\textwidth]{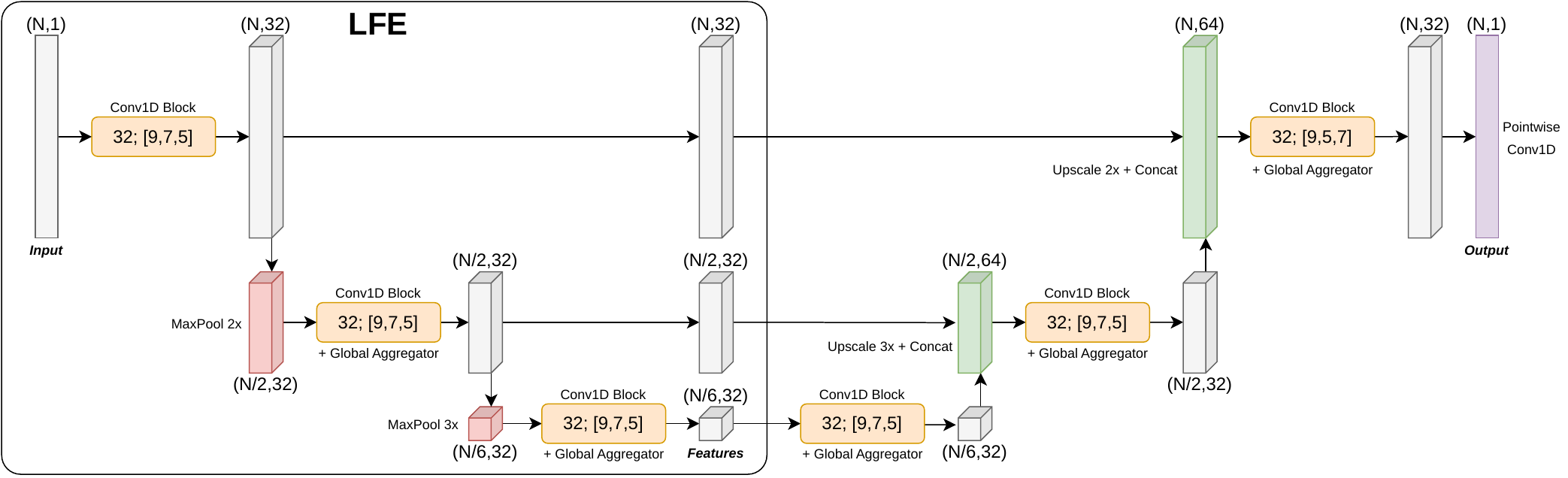}
    \caption{
        Laser Feature Extractor (LFE) network architecture, applied to a segmentation task.
        Each 1D convolutional block consists of three consecutive depthwise separable \cite{separable2017}
        1D convolutions of different kernel sizes (9, 7 and 5 respectively).
        Some blocks also contain a global feature aggregator, which performs a global maxpool
        of the input and concatenates the resulting features to each individual position of the input.
        Finally, a residual path adds the input of the block to the output of the last convolution.
        The segmentation mask is generated by an ``inverse'' LFE similar to U-Net \cite{unet2015}
        followed by a pointwise convolution that produces the final output logits.
    }
    \label{fig:lfe}
\end{figure*}

A 2D laser scan reading is usually presented as a 1D vector of range measurements,
the position of each element within which determining the angle of the laser beam
with respect to the origin. Deep learning algorithms learn to extract a set of
abstract features about their input
(as opposed to a specific set of features designed by humans),
and use those features to solve a given problem (such as classification).

We propose a new Fully Convolutional Network, called the Laser Feature Extractor (LFE),
which extracts features from a 1D vector of range measurements.
This network is inspired by image classification and segmentation networks such as
U-Net \cite{unet2015}, ResNets \cite{resnet2016} or MobileNet \cite{mobilenet2017}.
Its architecture (shown in Fig.~\ref{fig:lfe})
consists of a stack of residual/convolutional and maxpool downscaling layers used to
extract a feature map. There are three residual blocks in total, each containing
three convolutional layers. In order to reduce the search space and improve runtime
speed, all convolutional layers are depthwise separable \cite{separable2017}:
this means they are decomposed into two steps: a stack of independent convolutions
(one applied to each corresponding channel), followed by a pointwise convolution.
The activation function used after each convolution is ReLU, followed by batch
normalization and dropout layers.

LFE generates features at three different levels of downsampling: the original resolution
of the range data, the data downsampled by 2, and the data downsampled by 6 (in other words, combined downsampling by 2 and 3). This is
especially useful given the polar nature of range data, thus allowing features to be
extracted at close distances (where the input resolution is bigger) and also at farther
distances (where the input resolution is smaller). The downsampling factors have been chosen to increase the likelihood of evenly dividing the number of points in the laser scan vector (for instance, DROW's 450 points cannot be divided by 4).

\subsubsection{Training protocol} \label{sect:lfetrain}

While the backbone of an object detector is traditionally pretrained with a simpler
classification problem (and dataset such as ImageNet \cite{imagenet2009}), there is no such
equivalent available for 2D laser scan data. In order to validate LFE on its own, we consider the
laser scan segmentation problem (similar to PeTra \cite{petra2019}), and use it to allow
LFE to learn relevant features for detecting people. In order to use LFE in a
segmentation problem, we attach an ``inverse LFE'', making the whole network similar to
U-Net \cite{unet2015}. This network thus learns a binary label for each input point,
identifying which points are part of people's legs (and which are not). The segmentation output can also be post-processed with classical algorithms (such as
SciPy's \texttt{find\_peaks} function) in order to generate discrete people detections,
something which we will revisit in a later section of this paper.

\subsection{People Proposal Network (PPN)}

\begin{figure*}
    \centering
    \insetfont
    \includegraphics[width=\textwidth]{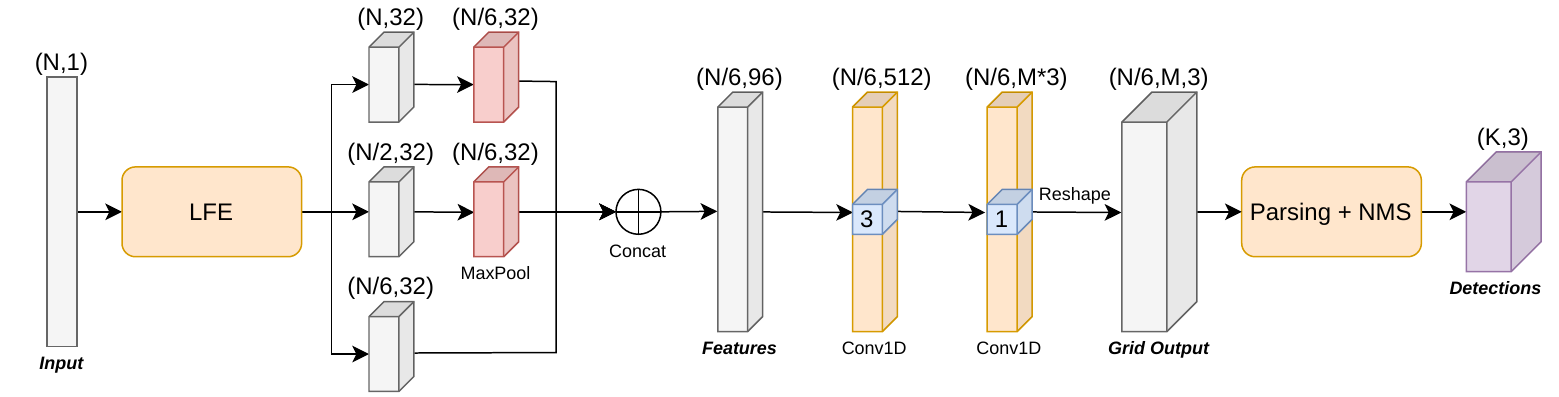}
    \caption{
        People Proposal Network (PPN) architecture, incorporating a LFE backbone.
        The features extracted by the LFE are further processed by a depthwise
        separable \cite{separable2017} 1D convolutional
        layer with kernel size 3, after which the outputs ($M \times 3$) for each
        sector in the grid are generated by a final pointwise 1D convolutional layer. The
        Non Maximum Suppression (NMS) process parses the grid output and generates the final
        people detections.
    }
    \label{fig:ppn}
\end{figure*}

\begin{figure*}
    \centering
    \insetfont
    \includegraphics[width=0.8\textwidth]{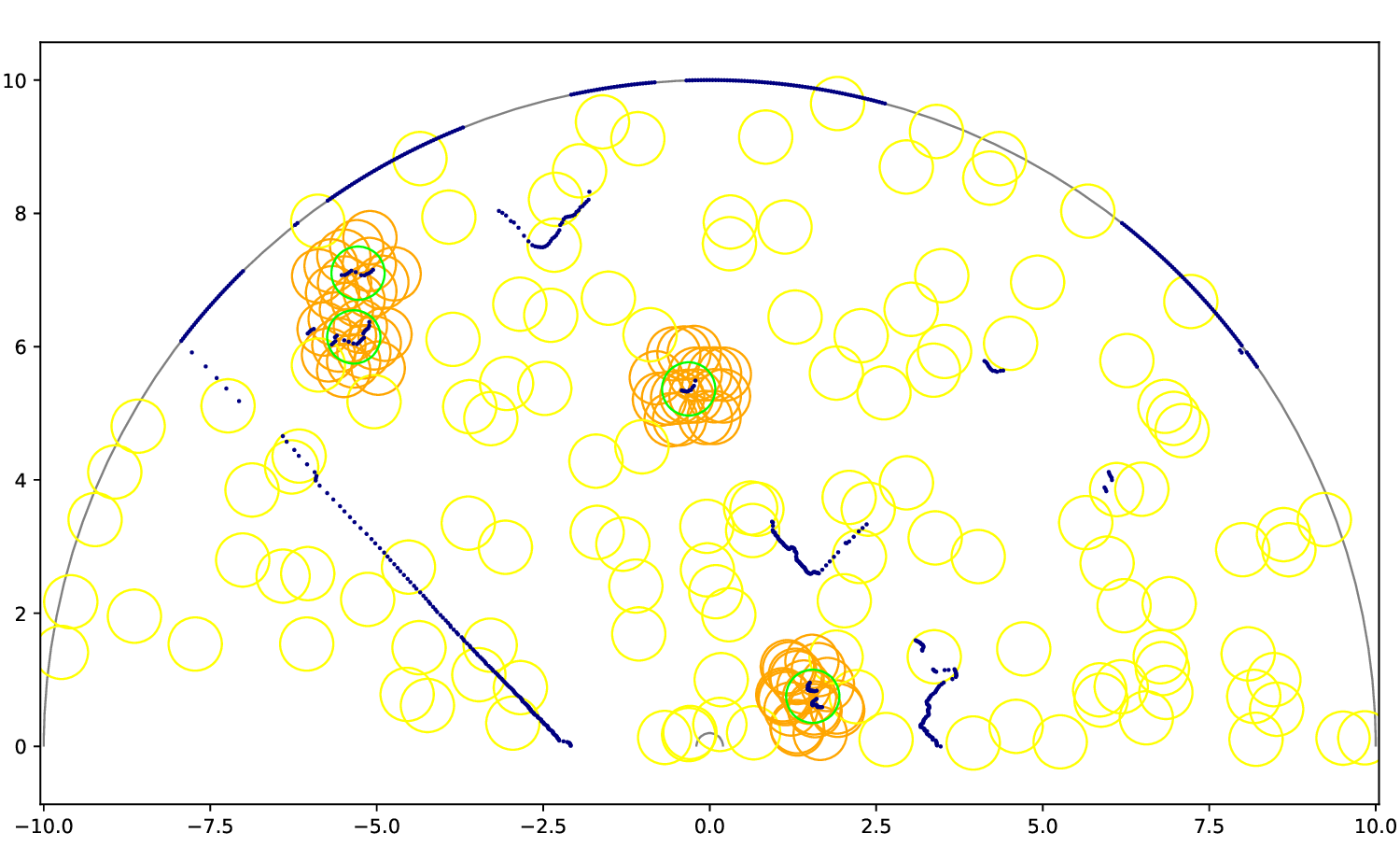}
    \caption{Example anchor grid used by the People Proposal Network
    during training on the FROG dataset.
    The boundaries of the field of view are drawn in gray.
    The laser scan data is represented by navy blue dots.
    Green circles are ground truth people annotations.
    Orange and yellow circles are examples of positive and negative anchor circles respectively (not all shown).
    An anchor is considered as positive ($s=1$) if it is close enough to a ground truth person.}
    \label{fig:froggrid}
\end{figure*}

The second component of the network is the People Proposal Network (PPN). This network
(shown in Fig.~\ref{fig:ppn})
is directly inspired and based on the Region Proposal Network (RPN) introduced by
the foundational object detection work Faster-RCNN \cite{fasterrcnn2015}.
We adapt the anchor grid of object proposals of the RPN so that it can be used in
the fundamentally 1D problem of laser scan data people detection.
In the original RPN the anchor grid is bidimensional; each element of the
feature map represents a 2D subarea of the original image, and several anchors
are trained in parallel for each 2D subarea using different aspect ratio priors.
In our People Proposal Network the anchor grid corresponds to sectors of the full
field of view of the laser, with their amplitude and number determined by the
largest downsampling performed by LFE.
We place multiple anchors at each sector, each having different
distance priors fully and evenly covering the entire range of the depth axis.
The field of view of the PPN has the shape of a circular ring sector
(see Figure~\ref{fig:froggrid}).
An important consequence of this design is that anchors are more densely placed
in central (near) areas of the field, while being sparser at far areas. This
is due to dealing with a polar coordinate system as opposed to a Cartesian
coordinate system -- this is in line with the nature of 1D laser scan data.

\begin{table}[b]
    \centering
    \insetfont
    \caption{
        Statistical information (mean and standard deviation)
        about the two regression targets in the generated training data:
        distance offset ($\Delta d$) and arc offset ($\Delta \ell$).
    }
    \begin{tabular}{l c c}
        \toprule
        \multicolumn{1}{c}{} & $\mu$ & $\sigma$ \\
        \midrule
        $\Delta d$    & 0.0605 & 0.9166 \\
        $\Delta \ell$ & 0.0019 & 0.8773 \\
        \botrule
    \end{tabular}
    \label{table:regr}
\end{table}

The PPN receives the feature maps extracted by LFE as input,
and outputs three target values for each anchor:
$s$ (objectness), $\Delta d$ (distance offset) and $\Delta \ell$ (arc offset).
The feature maps generated at different downsampling levels are maxpooled into the
same resolution and concatenated in order to generate a single input feature map.
The objectness score is learned as a classification problem, while the
distance/arc offsets are learned as a regression problem. These offsets
are centered on each individual anchor's center point.
We learn the arc offset instead of the angle offset so that the
learned offsets are in the same displacement scale as the distance offsets,
instead of having drastically different scales depending on how close to the origin
each individual anchor is. Furthermore, we normalize their scale by dividing both
offsets by the anchor spacing along the depth axis: $\frac{d_{\text{far}} - d_{\text{near}}}{N_{\text{anchors}}}$, where $d_{\text{near}}$ and $d_{\text{far}}$ are the minimum and maximum detection distances considered, respectively.
We also empirically show that resulting target offsets used for learning
have a close-to-normal distribution
($\mu_{\Delta d}, \mu_{\Delta \ell} \approx 0,
\sigma_{\Delta d}, \sigma_{\Delta \ell} \approx 1$), see Table~\ref{table:regr}.

At inference time, the distance and arc offsets of each anchor are decoded into Cartesian XY coordinates representing the center of a bounding circle, and paired with their corresponding classification scores. Note that these centers are usually located between the two legs of a person, and they do not necessarily correspond to individual sensor measurements; in fact they rarely do (if ever). All circles are defined to have the same radius.
Like \cite{fasterrcnn2015}, the output from the network then undergoes a Non-Maximum
Suppression (NMS) filter. Traditional NMS as applied in object
detection is based on the Intersection over Union (IoU) measurement between
bounding box proposals.
In our case we use a simple distance function between person center
proposals. In other words, two person proposals overlap if the distance between
their centers is smaller than a given hyperparameter, which usually matches the
most common ground truth circle diameter.

\subsubsection{Training protocol}

The training process of the network involves generating anchor classification
and regression data for every scan, based on the person annotations in the ground
truth. In a similar way to \cite{fasterrcnn2015}, we group all anchors in the grid
into two categories: positive and negative. The overlap metric
used as criterion is once again the distance between circle centers, and the
boundary between groups is a tunable hyperparameter.

The loss function used is the following:

\begin{equation}
    \mathcal{L} = \frac{1}{N_{+} + N_{-}} \mathcal{L}_{cls} + \frac{1}{N_{+}} \mathcal{L}_{reg}
\end{equation}

\noindent where
$\mathcal{L}_{cls}$ is the classification loss,
$\mathcal{L}_{reg}$ is the Smooth L1 regression loss and
$N_{+}$/$N_{-}$ are the number of positive/negative anchors respectively.
Positive and negative anchors both contribute to classification loss, while only
positive anchors contribute to regression loss. Both losses are scaled so that
they have the same magnitude, by virtue of dividing by the number of contributing
anchors respectively.
Given the large imbalance between
positive and negative anchors, we combine the traditional
binary cross-entropy loss with the Dice loss \cite{diceloss2017} (a smooth variant
of the $F_1$ score), taking the average of the two \cite{comboloss2023} as classification loss.
We do not follow the approach in \cite{fasterrcnn2015}
(resampling the positive and negative anchors within each scan to be in a
fixed 1:1 proportion, with each scan producing the same overall number of anchors)
because said imbalance is greater than the one in 2D object detection,
causing that approach to be rendered unfeasible.
This is also an effect of the unevenness of anchor density with respect to distance,
meaning a large number of scans do not contain enough positive anchors to properly
fill the desired quota.

\section{FROG benchmark} \label{sec:bench}

We propose using the FROG dataset as a new benchmark for 2D laser range finder based
people detectors.
As such, we carry out several experiments with existing
detectors, as well as our own proposed detectors. In particular, we select
the DROW3 \cite{Beyer2018RAL}, DR-SPAAM \cite{Jia2020DRSPAAM} and PeTra \cite{petra2019} detectors from the state of the art for an initial
benchmark based on the FROG dataset, in addition to a well known baseline provided by the ROS framework \cite{RosLeg2010R}.
The benchmark codebase we developed to perform these experiments can be found on
GitHub\footnote{\url{https://github.com/robotics-upo/2DLaserPeopleBenchmark}},
and it provides a common implementation of all metrics and evaluation protocols
for maximum consistency.

We define a subset of the FROG dataset to be used in this benchmark, containing training/validation and testing sets.
The training/validation set is sourced from two different sequences recorded around the time of greatest attendance (around noon, maximizing the number of person annotations) and later randomly split in 90:10 proportion.
The testing set is sourced from another different sequence.
In both cases, scans with empty lists of person annotations are excluded from the benchmark. Models are trained on the training set, and metrics are calculated and reported on the testing set. The validation set is only used to provide feedback during the training process, as well as optimizing hyperparameters.
We provide all the data, and do not withhold the labels associated with the testing set.

\subsection{Evaluation criteria and process} \label{sec:met}

We follow existing practices in
\cite{BeyerHermans2016RAL,Beyer2018RAL,Jia2020DRSPAAM,Jia2021Person2DRange},
and use the same metrics for evaluation purposes.
These metrics revolve around the Precision-Recall (PR) curve, which is intended to show
the overall performance profile of the model at different desired precision/recall tradeoffs.
Particularly, we consider the following:

\begin{itemize}
\item
    \textbf{Average Precision}: This is the main evaluation metric used by the
    object detection community, and it is nominally equivalent to the area under the
    PR curve (AuC). However, estimating this area can be a challenging process due to
    discontinuities created by small variations in example ranking. For this reason,
    we follow the object detection community (specifically MS COCO {\cite{mscoco}})
    in using the 101-recall-point interpolation method to calculate this metric.
    This contrasts with \cite{BeyerHermans2016RAL}, which applied the trapezoidal rule instead.
    As a note, we believe this metric produces unexpected behavior when evaluating certain methods. In Section~{\ref{results}} this is explained with more detail.
\item
    \textbf{Peak\,F1} score: This is the maximum F1 score obtained along the PR curve.
    Note that the F1 score is the harmonic mean between the Precision and Recall values.
\item
    \textbf{Equal Error Rate} (EER): This is the closest value along the PR curve at which
    Precision equals Recall.
\end{itemize}

These metrics can be parametrized: for example, AP$_{d}$ considers detections to be positive
if there exists an unmatched ground truth annotation within $d$\,m of the detection.
This $d$ parameter is known as the \textit{association distance}, and it is analogous to the IoU threshold of object detection metrics.
Note that only people centers are considered, as opposed to full circles.
Following established 2D LiDAR people detection work \cite{Beyer2018RAL,Jia2020DRSPAAM}, we calculate the PR curve and evaluate all associated metrics
using two different values for $d$: 0.5\,m and 0.3\,m.
In addition, we calculate averaged mAP, mPeak\,F1 and mEER metrics for $d = [0.3:0.05:0.5]\,\text{m}$ in order to capture overall performance at different association distances with a single value, similarly to MS COCO and its mAP metric over a range of IoU thresholds.

In order to calculate the PR curve, we first obtain the collection of person detections
produced by each model for each scan in the tested split. Each person detection is expected
to have a confidence score (0.0 to 1.0), and X/Y positions. We ignore very low
confidence detections (with score $<$ 0.01), as well as detections falling outside a 10m
distance threshold; so that the evaluation process takes a reasonable amount of time,
and also to avoid unfairly rewarding methods that generate hundreds of noisy detections
with very low confidence scores.
Within each scan, detections are matched to their closest ground truth annotation.
If multiple detections are matched to the same ground truth, the one closest to the
annotation becomes a True Positive and the others are considered False Positives.
Detections that are farther than $d$ to any ground truth annotation are also considered
False Positives.
Following existing aforementioned works, the matchups between detections and ground truth circles are recalculated after processing every detection (in descending order of confidence score), so that higher confidence
detections can initially count as True Positives in case a lower confidence detection (processed later) is closer to a ground
truth annotation -- this ultimately only replaces one detection for another, and does not change the total tally of True Positives.
A final global aggregation process traverses the complete list of detections across scans
(also in descending order), accumulating a count of True Positives and thus computing
every ${(P,R)}$ point of the curve. If multiple detections share the same confidence score,
they are summarized as a single point in the PR curve.

Lastly, we also report the end-to-end inference time as an evaluation metric. Specifically, we use the ROS environment with each detector's provided ROS node, and measure the total time taken by the node to process each laser scan and output detected people. This is calculated by playing back the test set at a laser frequency high enough to bottleneck all detectors, and dividing the simulated time by the number of received people detection messages. Concretely, we play back the test set at 20 times the speed, resulting in a 800Hz laser frequency sustained for 88.49s.

\subsection{Experimental setup}

\begin{table*}[ht!]
    \insetfont
    \caption{
    Benchmark results for several person detector models. Besides the baseline, all models are
    trained on FROG's \texttt{train} split, and evaluated on FROG's \texttt{test} split.
    Average Precision (AP), Peak $F_1$ and Equal Error Rate (EER)
    are reported for two association distances: 0.5\,m and 0.3\,m, as well as averaged across $[0.3:0.05:0.5]\,\text{m}$.
    End-to-end inference times obtained with each detector's ROS node are also reported, in milliseconds.
    }
    \begin{tabular}{l c c c c c c c c c c}
        \toprule
        \multicolumn{1}{c}{} & \multicolumn{3}{c}{} & \multicolumn{3}{c}{d = 0.5\,m} & \multicolumn{3}{c}{d = 0.3\,m} & \multicolumn{1}{c}{} \\
        \cmidrule{2-4}\cmidrule{5-7}\cmidrule{8-10}
        \multicolumn{1}{c}{} & mAP & mPeak\,$F_1$ & mEER & AP & Peak\,$F_1$ & EER & AP & Peak\,$F_1$ & EER & Time (ms)\\
        \midrule
        ROS \texttt{leg\_detector} & 15.8 & 30.8 & 30.5 & 20.2 & 35.2 & 34.9 & 10.0 & 24.3 & 24.1 & 1.77 \\
        \midrule
        PeTra               & 49.6 & 66.4 & 66.2 & 50.1 & 66.6 & 66.4 & 49.1 & 66.1 & 65.9 & 28.17 \\
        PeTra*              & 58.3 & 67.7 & 67.1 & 59.0 & 67.9 & 67.3 & 57.9 & 67.4 & 66.8 & ---   \\
        LFE-Peaks (ours)    & 64.9 & 70.2 & 70.2 & 65.6 & 70.7 & 70.7 & 63.2 & 69.0 & 69.0 &  1.76 \\
        LFE-PPN (ours)      & 66.5 & 68.7 & 68.6 & 69.2 & 69.5 & 69.5 & 62.5 & 67.3 & 67.2 &  1.49 \\
        DROW3 $(T = 1)$     & 73.6 & 71.9 & 71.6 & 73.9 & 72.0 & 71.8 & 73.0 & 71.6 & 71.4 & 13.08 \\
        DR-SPAAM $(T = 1)$  & 73.3 & 71.9 & 71.6 & 73.7 & 72.1 & 71.8 & 72.7 & 71.6 & 71.3 & 13.95 \\
        DR-SPAAM $(T = 5)$  & 75.3 & 73.4 & 73.3 & 75.6 & 73.6 & 73.4 & 74.7 & 73.2 & 73.0 & 13.99 \\
        \botrule
    \end{tabular}
    \label{table:benchmarkresults}
\end{table*}

\begin{figure*}[ht!]
    \centering
    \insetfont
    \includegraphics[width=0.49\textwidth]{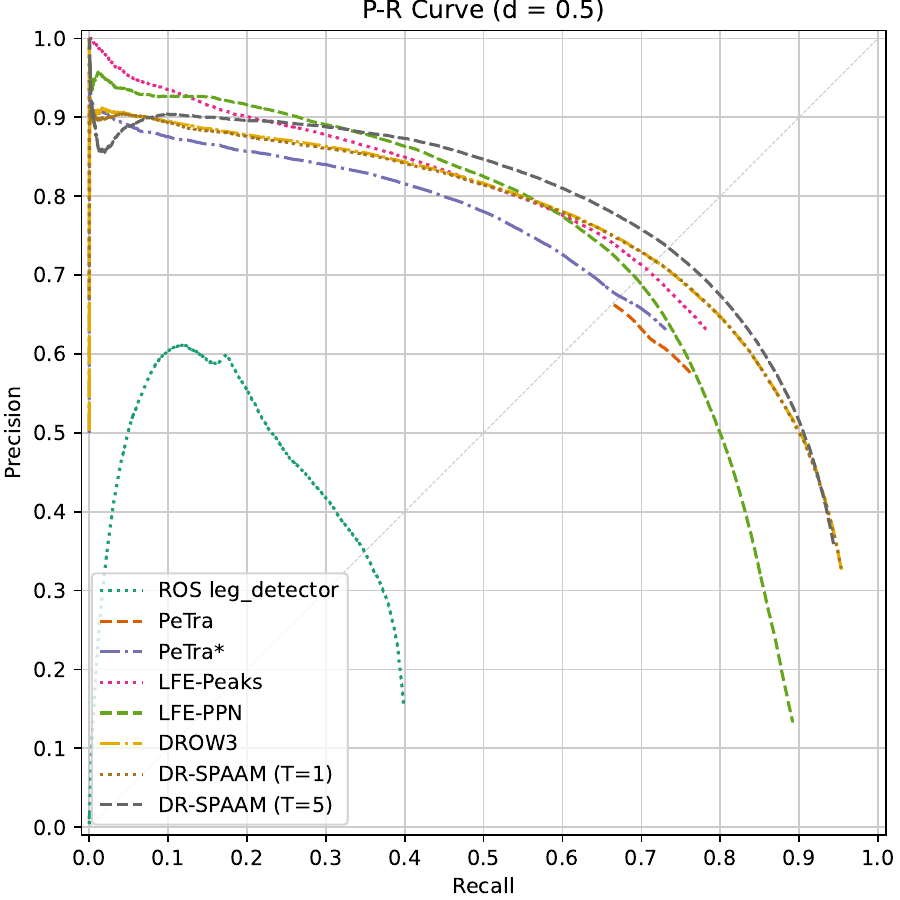}
    \includegraphics[width=0.49\textwidth]{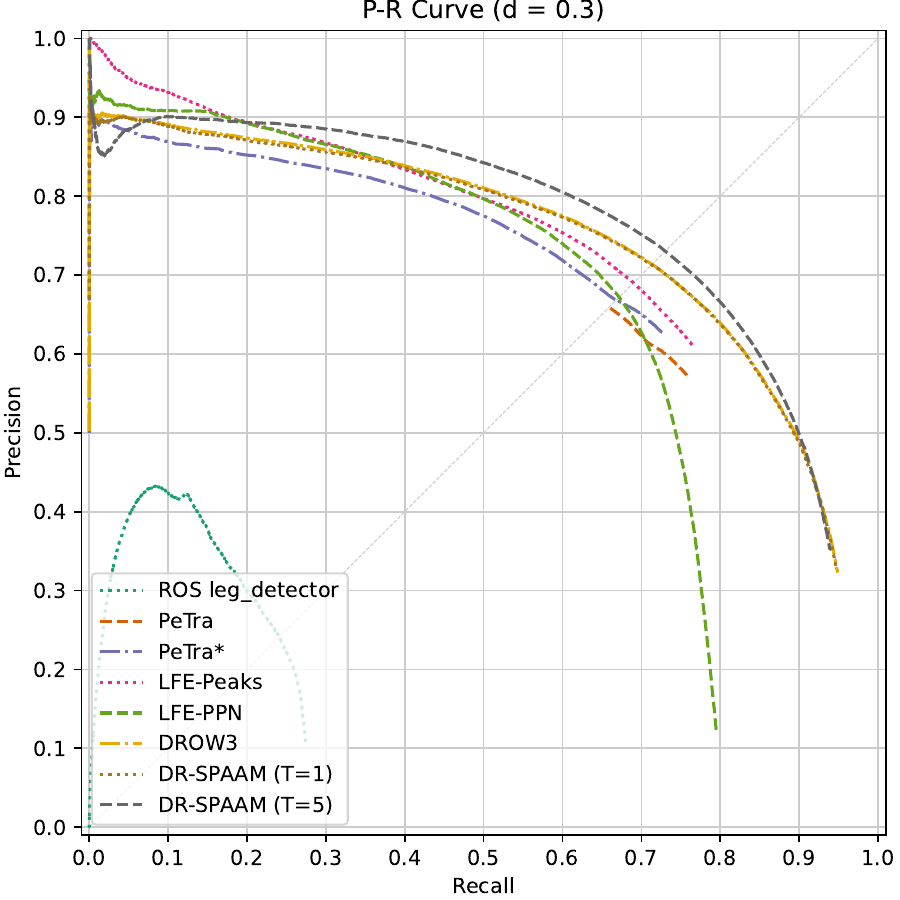}
    \caption{
        Precision-recall curves for the person detector models. We show
        curves for both association distances: 0.5\,m and 0.3\,m.
    }
    \label{fig:benchmark}
\end{figure*}

\begin{figure*}[ht!]
    \centering
    \insetfont
    \begin{tabular}{@{}c@{}c@{}}
    \includegraphics[width=0.49\textwidth]{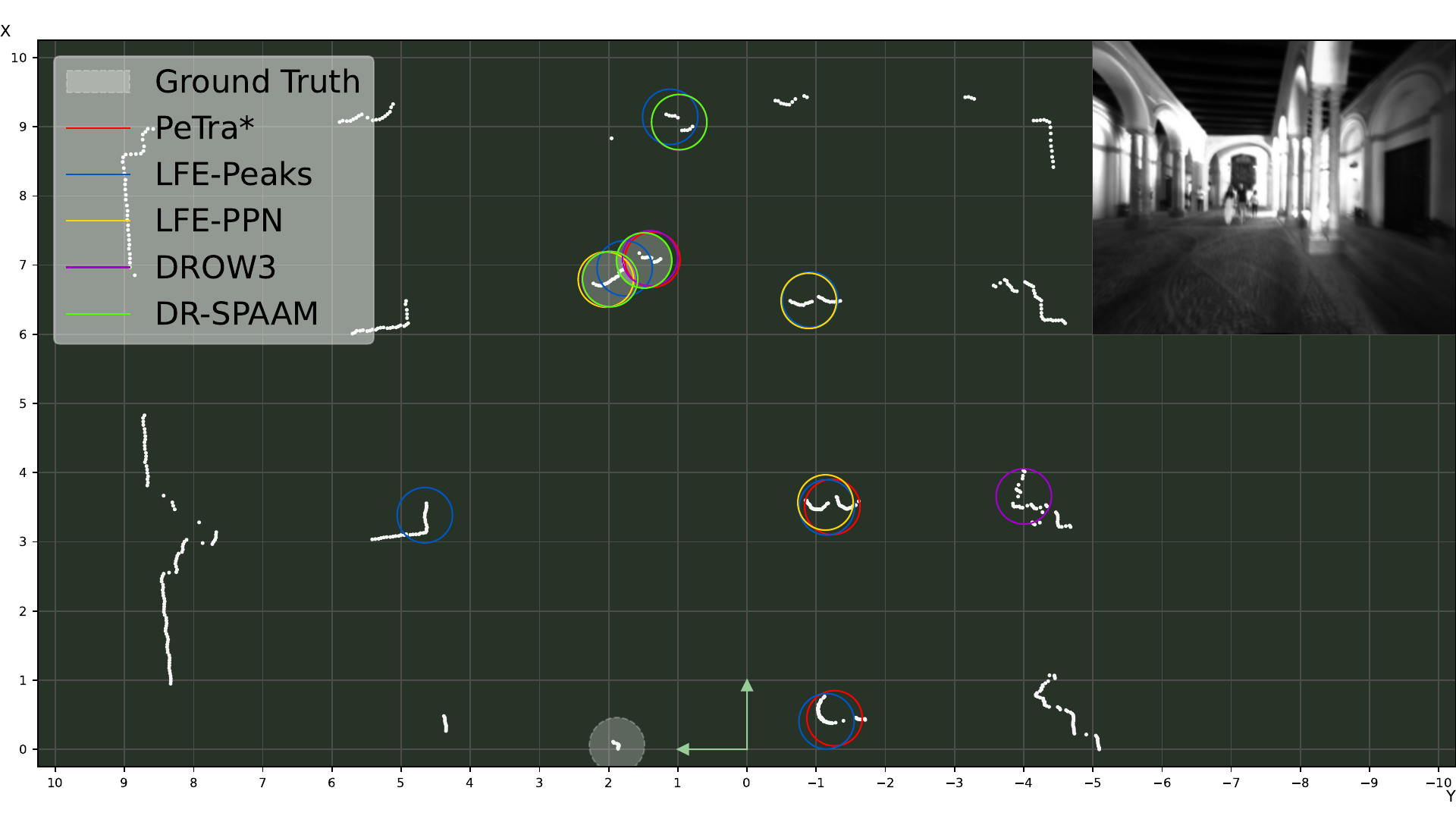} &
    \includegraphics[width=0.49\textwidth]{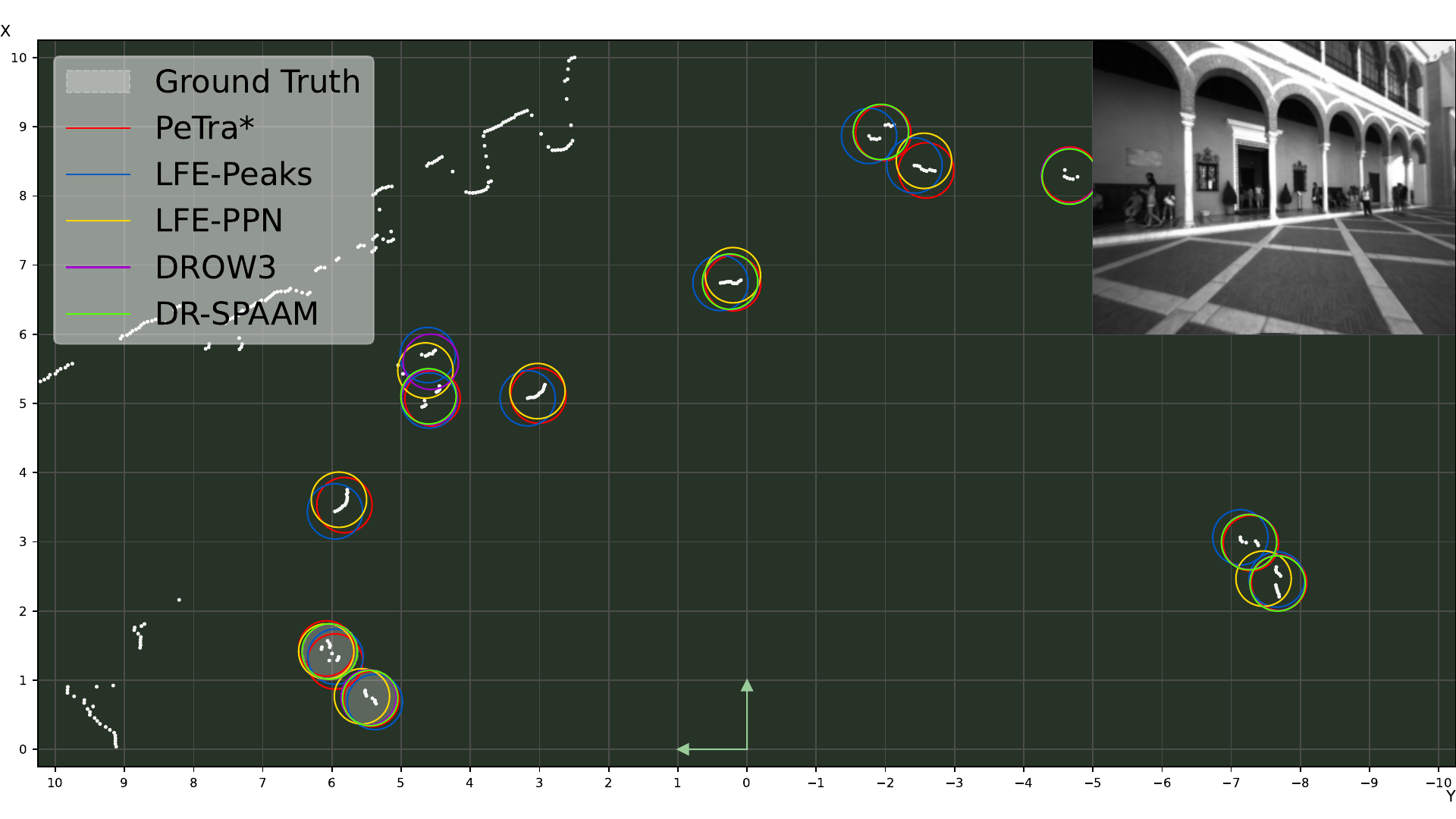} \\
    Scene 1 & Scene 2 \\
    \includegraphics[width=0.49\textwidth]{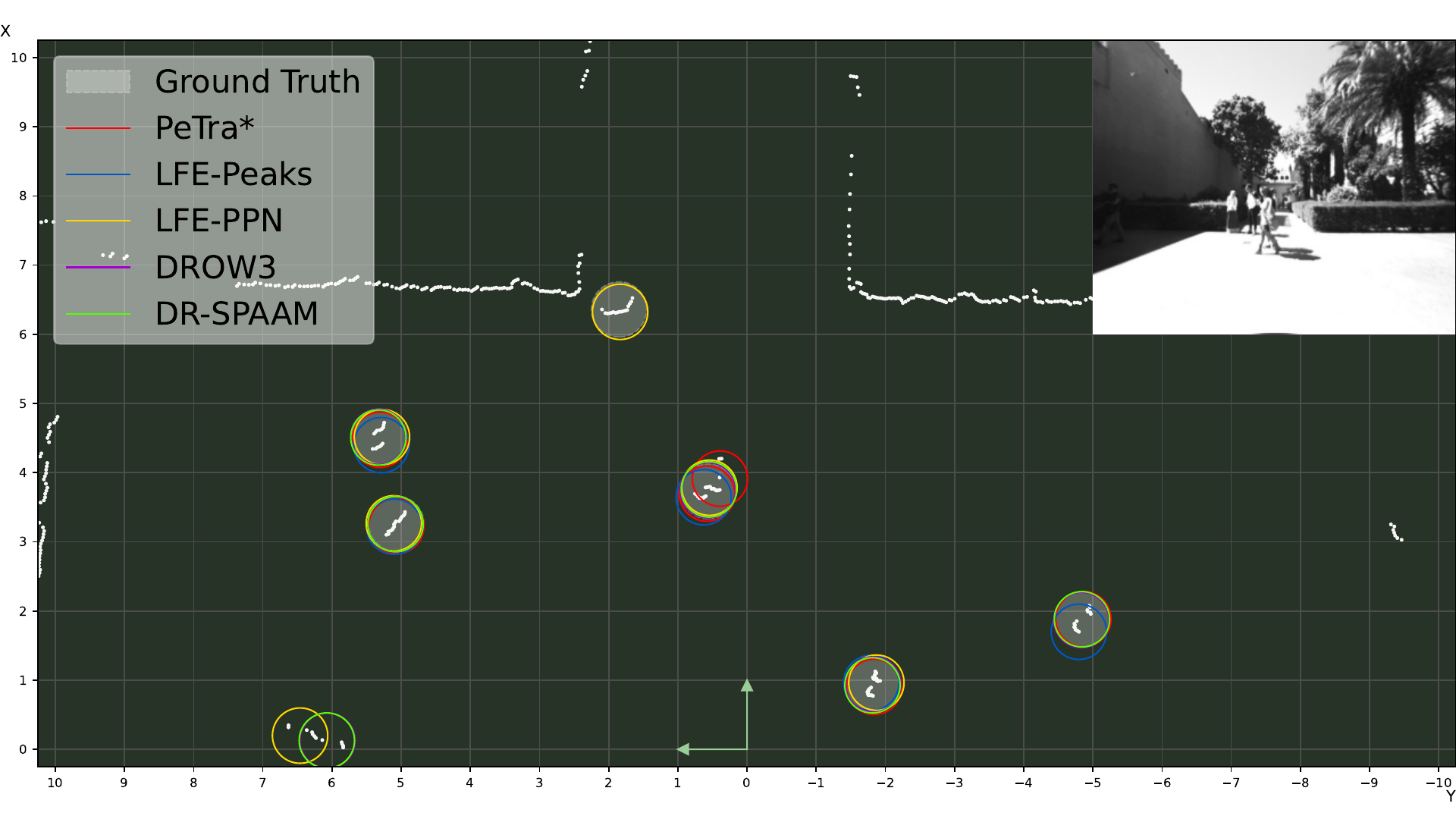} &
    \includegraphics[width=0.49\textwidth]{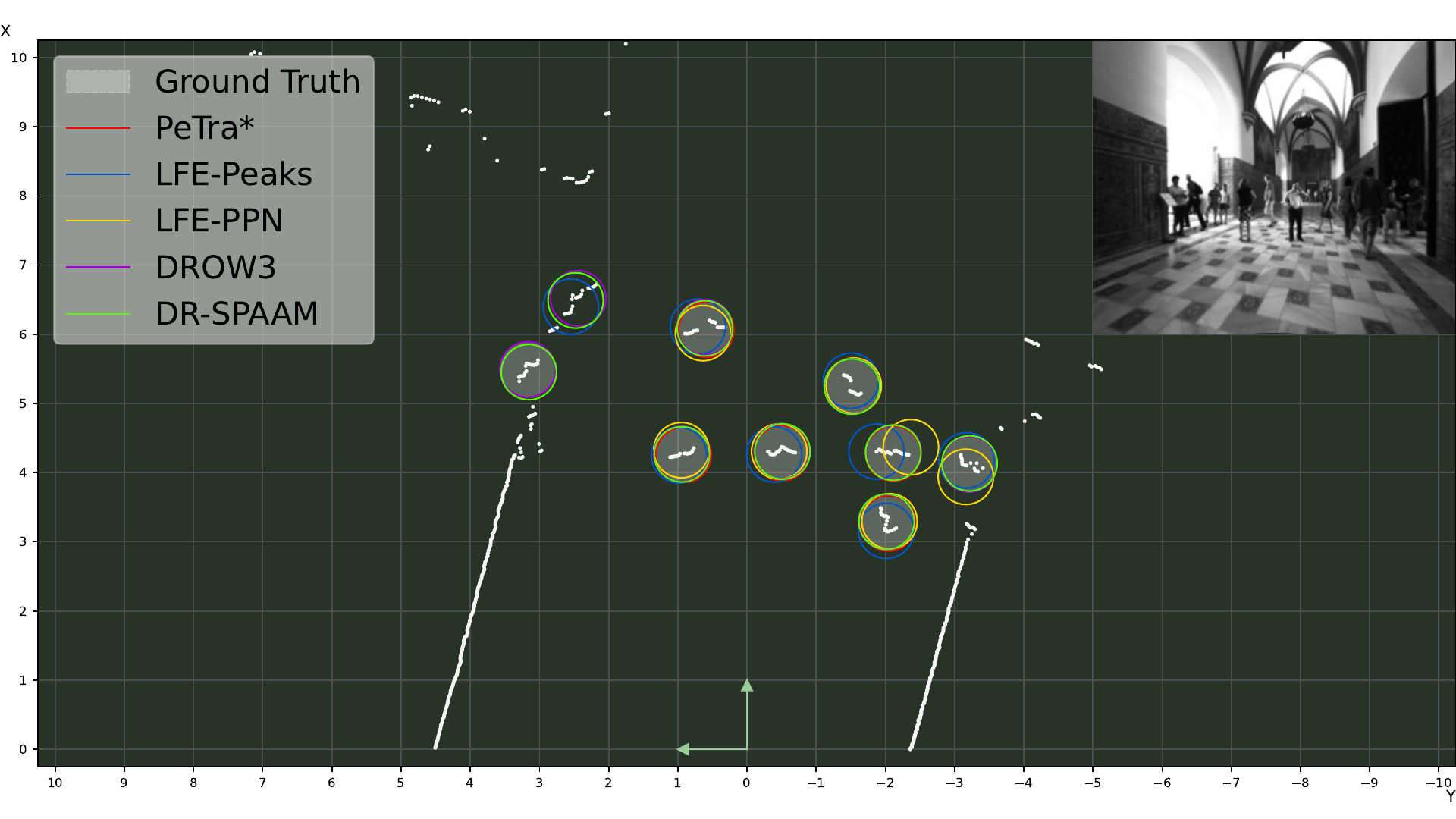} \\
    Scene 3 & Scene 4 \\
    \includegraphics[width=0.49\textwidth]{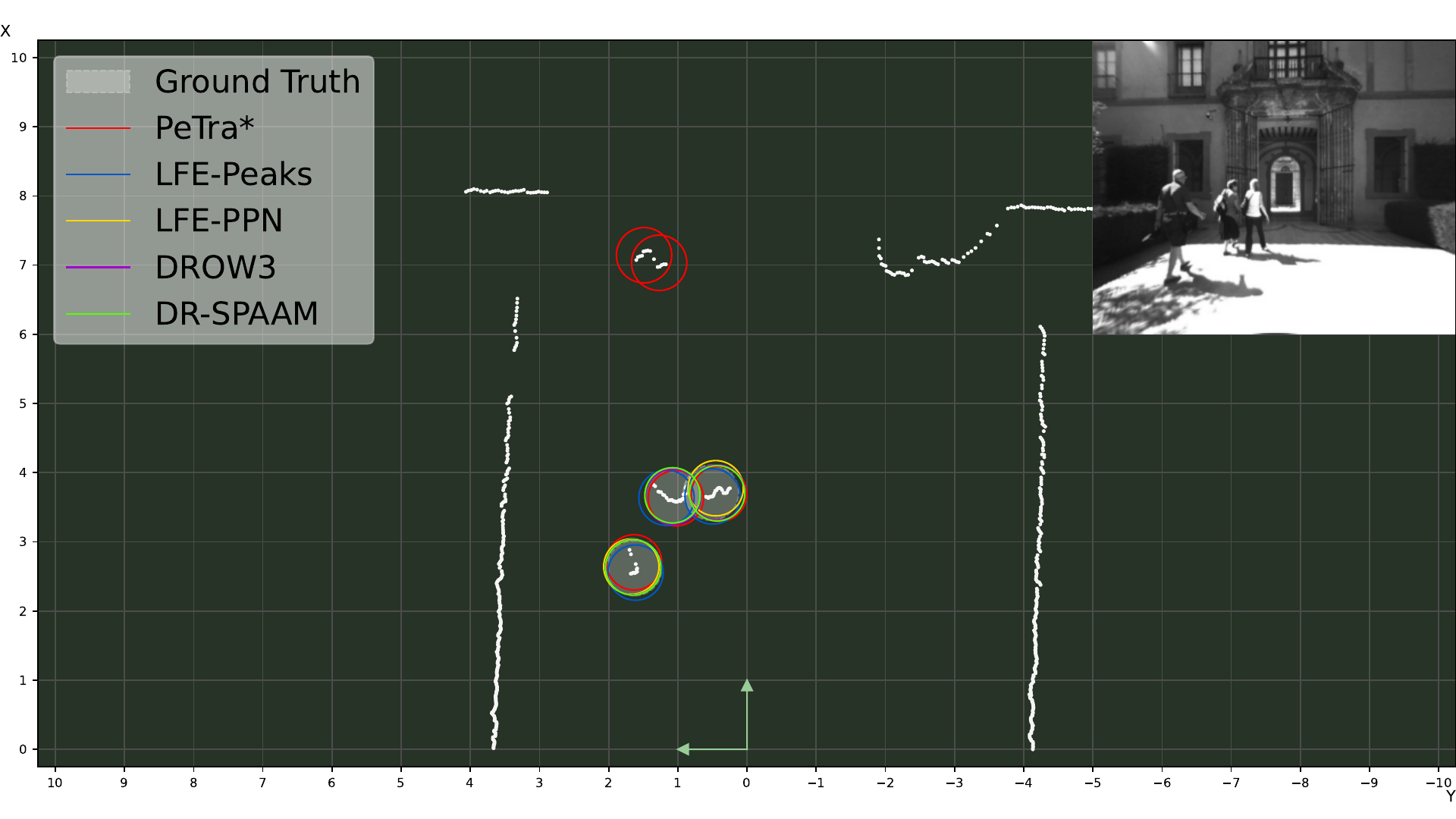} &
    \includegraphics[width=0.49\textwidth]{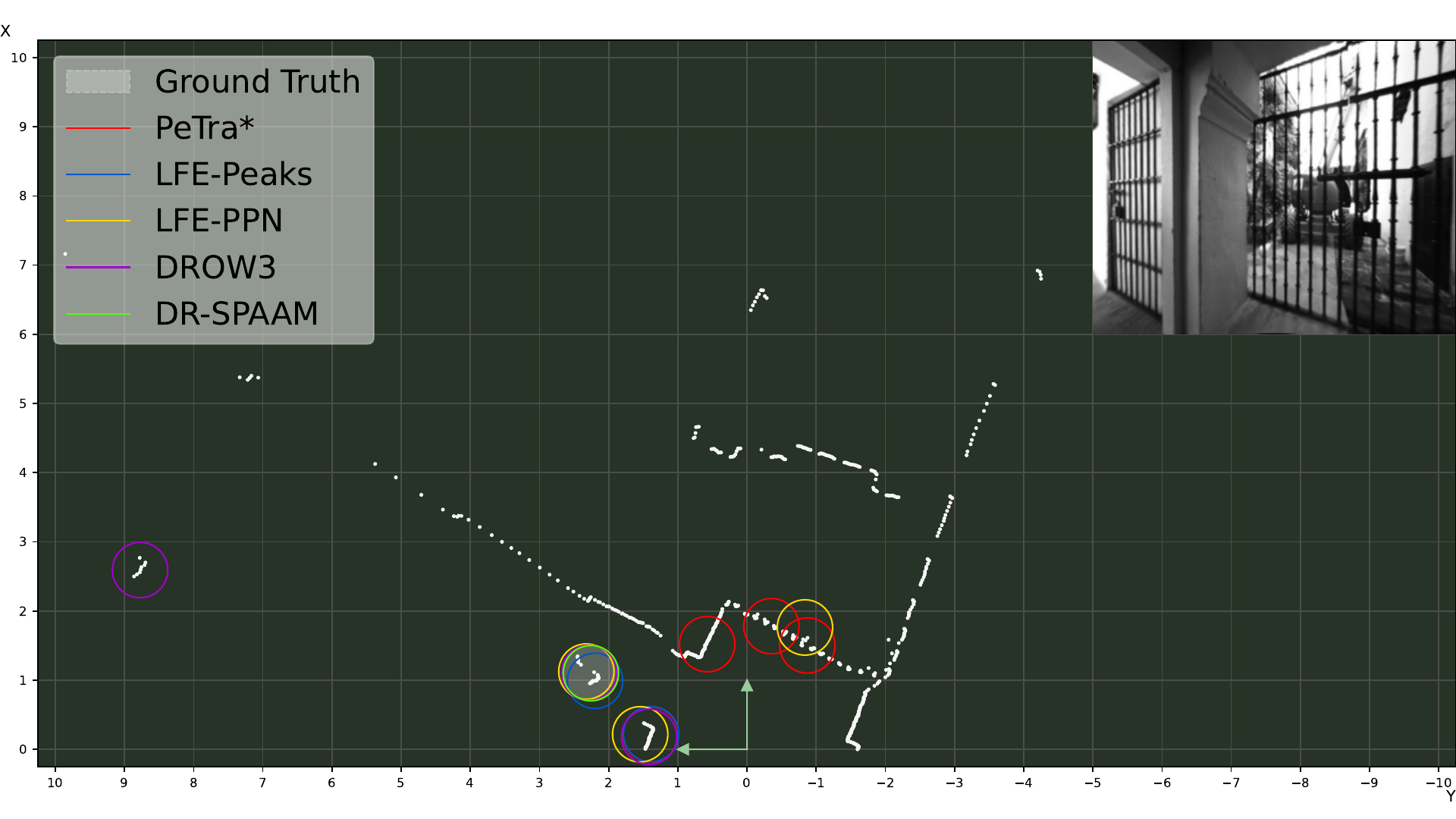} \\
    Scene 5 & Scene 6 \\
    \end{tabular}
    \caption{
    Collection of qualitative results, sampled from the FROG dataset's
    \texttt{test} set. In the case of PeTra and DR-SPAAM, we only show
    the best variant (PeTra* and DR-SPAAM with T = 5) for clarity.
    Several scenes showcasing the performance of the detectors in
    different types of environments are included.
    }
    \label{fig:qualit}
\end{figure*}

As mentioned previously, we select a baseline and several state-of-the-art detectors (as well as our own
detectors) to be evaluated as part of the first benchmark using the FROG dataset.
In this section we will discuss the exact methodology used to test each detector,
including hyperparameters used, challenges encountered during evaluation, adaptations
needed to produce meaningful results, and other miscellaneous details.

The platform used to train and evaluate models is a desktop PC sporting an Intel Core i9-9900X CPU with 128 GB of RAM and an NVIDIA TITAN RTX GPU with 24 GB of VRAM.

\subsubsection{ROS \texttt{leg\_detector} baseline}

ROS 1 distributions offer a standard package containing a pre-trained 2D laser based leg detector/person tracker. This detector implements a variant of \cite{Arras2007UsingBF}, a classical algorithm based on hand-crafted geometric segment features followed by a random forest classifier. We include its results in the benchmark because it is a commonly used solution in the field of robotics, making it relevant as a baseline to which compare the performance of other detectors.
In order to more accurately represent the baseline as typically employed in robotics projects, we follow \cite{Beyer2018RAL} and use the provided pre-trained forest as-is (with all default hyperparameters); i.e. not retraining the model with data from the FROG dataset. People tracking measurement results are captured via ROS bag, and later converted into the common evaluation format expected by our benchmarking codebase.

\subsubsection{PeTra}

We evaluate PeTra \cite{petra2019} as a modern representative of leg-based detectors. This work
essentially replaces ROS \texttt{leg\_detector}'s classical
algorithm with an image-based 2D fully convolutional segmentation network, which detects
points belonging to people's legs. This segmentation output is later post-processed
in order to extract individual leg locations, and pairs detected legs to produce
final person location proposals. The source code was made available online by
its authors\footnote{\url{https://github.com/ClaudiaAlvarezAparicio/petra}}.

We train PeTra's segmentation network using $512 \times 256$ as the resolution of
the 2D image onto which the laser scan data is projected, with the height dimension
being half of the width dimension in order to save computational resources
(given that the FoV of the laser is $180^\circ$, and otherwise the bottom half of
the image would always remain unused).
In addition, we randomly sample a set of
12000 scans from FROG's training dataset to convert into projected 2D images instead of converting them all,
due to the increased memory and computation requirements of working with 2D images
as opposed to 1D laser scan vectors.

We train the network for 30 epochs with a batch size of 16 scans using
the Adam optimizer with $\eta = 5 \times 10^{-4}$.

PeTra uses OpenCV's \texttt{findContours} routine to find shapes corresponding to
legs; however it does not generate a detection score. In order to solve this, we take the
average output of the network associated with each point belonging to each
contour as its overall detection score. Unfortunately, due to the fact that PeTra
is trained using Dice loss, many points end up with the maximal 1.0 score
allowed by the output of the sigmoid (after the inevitable loss of precision),
meaning that a large number of detections are generated with confidence 1.0.
This results in a very short PR curve, as can be seen in Fig.~\ref{fig:benchmark}.
To remedy this problem, we also evaluate PeTra with the same mixed loss
function used by our own segmentation-based detector, resulting in a model we
called \textbf{PeTra*}. This allows us to lower the overconfidence of the
detections, and thus plot a more meaningful PR curve for PeTra.

We perform inference speed tests using PeTra's provided Docker image based on ROS 1 Melodic. We do not evaluate PeTra* separately because it only differs in model weights. In order to ensure accurate results, we set the subscriber queue length to 1, and move the detector code into the subscriber callback (the original code performs detections as a separate timer task, independent of laser frequency).

\subsubsection{DROW3 \& DR-SPAAM}

We evaluate DROW3 \cite{Beyer2018RAL} and DR-SPAAM \cite{Jia2020DRSPAAM}
as representatives of the cutout-based input preprocessing approach.
In particular, we evaluate the most up to date implementation of both models
\footnote{\url{https://github.com/VisualComputingInstitute/2D_lidar_person_detection}} with minimal modifications to the code in order to add support for loading the FROG
dataset. These modifications include adding custom training configuration files and disabling the code that deletes
certain log folders -- this allows us to directly read the generated detections and run our
common benchmarking code, which is shared with all other models we have evaluated.

We train and evaluate both DROW3 and DR-SPAAM in single-scan mode, that is, no information from
previous scans is used to generate predictions. We do this in order to enable a
fair comparison with other models that do not make use of temporal information.
In addition, we also train and present results for DR-SPAAM using a temporal window size of 5 scans, so that said approach
is also represented in the benchmark.
DROW3 cannot be evaluated in multi-scan mode because the codebase used for training does not
support reading odometry information.
We use the same cutout hyperparameters selected in \cite{Jia2020DRSPAAM} for
optimal person detection, in particular: 56 points, \mbox{(1.0m $\times$ 1.0m)} window size.

We train the networks for 5 epochs with a batch size of 8, the Adam optimizer,
and with an exponential learning rate schedule
ranging from $10^{-3}$ initially to $10^{-6}$ at the end of training.
In particular, we use 5 epochs so that the overall number of minibatches processed during training
on the FROG dataset is of the same order of magnitude as when training using the DROW dataset.

We perform inference speed tests using ROS 1 Noetic, the provided \texttt{dr\_spaam\_ros} package, and the latest available version of PyTorch at the time of writing (v2.0.1 with CUDA support). ``DR-SPAAM*'' is not included in the comparison due to not being included in the ROS package provided by the authors.

\subsubsection{LFE \& PPN}

We first train LFE on the segmentation problem described in Section~\ref{sect:lfetrain},
using a batch size of 32 and
the AdamW \cite{adamw} optimizer with $\eta = 10^{-3}$ and $\lambda = 4 \times 10^{-3}$.
We set a target of 100 epochs,
with an early stopping patience of 20 epochs and $\Delta \mathcal{L}_{min} = 10^{-3}$.
As mentioned previously, we also evaluate LFE on its own as a person detector
by adding a classical post-processing step based around SciPy's \texttt{find\_peaks} function, a combination we are calling \textbf{LFE-Peaks}.
The post-processing finds the height and width
of the peaks in the segmentation signal, and afterwards computes centroids based on the Cartesian
coordinates of the corresponding points in the range data. Centroids that are close together
are interpreted as legs or part of legs, and merged together into final person detections
using a NMS-like process. Outlier points are discarded from each centroid, so that we do not
take into account parts of the background in the averaging formula.
In a similar way to our adaptation of PeTra, the confidence score is the average output
of the network corresponding to all points assigned to each detection.

LFE-PPN embeds LFE as its backbone, which can either be trained from scratch, or its weights reused from the LFE segmentation experiment and further fine-tuned during LFE-PPN training. For the purposes of this experiment, we follow and report the latter approach, although we have not observed any quantitative differences between the two.
We employ a batch size of 4,
use the AdamW optimizer with $\eta = 10^{-4}$ and $\lambda = 4 \times 10^{-4}$, and set
a target of 150 epochs with an early stopping patience of 20 epochs and $\Delta \mathcal{L}_{min} = 10^{-3}$.

We perform inference speed tests using ROS 2 Humble and our own developed implementation of a person detection node based on either LFE-Peaks or LFE-PPN, using C++ and ONNX Runtime v1.14.1\footnote{\url{https://onnxruntime.ai/}}
with the CUDA backend. In the case of LFE-Peaks, an embedded Python interpreter is used to execute the classical peak finding and post processing algorithm. Our ROS package is publicly available on GitHub\footnote{\url{https://github.com/robotics-upo/upo_laser_people_detector}}.

\subsection{Results} \label{results}

We present the quantitative results of the benchmark (shown in Table~\ref{table:benchmarkresults}), as well
as the corresponding Precision-Recall curves in Figure~\ref{fig:benchmark}.
Methods based on cutout preprocessing (DROW3 and DR-SPAAM) obtain the overall best metrics,
as expected of the state of the art; while our own proposed methods (LFE and PPN) achieve
good results considering how they utilize fewer or no hand-crafted features in their design. The ROS \texttt{leg\_detector} baseline produces very poor results, clearly indicating the end of its usefulness after the development of much better detectors; however they are consistent with the results obtained by \cite{Beyer2018RAL} on the DROW dataset.

Counter-intuitive behavior can be observed in the behavior of the Average Precision metric.
We suspect this metric rewards methods with a longer PR curve, said length being a result of
incorporating a larger number of low confidence/quality guesses, which serendipitously inflates
the maximum recall score (at the expense of precision). Methods which do not generate such
guesses are unfairly punished. For this reason we believe the Peak-F1 and EER metrics present
a fairer comparison in practice.

We can observe that the PR curves and metrics of \textbf{DROW3} and \textbf{DR-SPAAM}
in single scan mode are
practically identical. We believe this to be related to the lack of temporal information
used during training, indicating that the contributions in \cite{Jia2020DRSPAAM} are not
focused on improving the baseline network architecture introduced by \cite{Beyer2018RAL}.
On the other hand, full DR-SPAAM with a temporal window size of 5 scans achieves the best
metrics overall, although its PR curve dips below that of DROW3 in the low recall section.

\textbf{PeTra} presents a challenging problem during evaluation. As explained earlier,
its choice of loss function generates a large number of detections with the maximum possible
confidence score (1.0), causing a significant portion of the data to be lumped together as a
single point, which in turn results in a degenerate PR curve. We correct this by changing
the loss function to incorporate binary crossentropy. The resulting model (\textbf{PeTra*})
does not present this problem, and thus achieves results more in line with the ones obtained
by other models. Despite being limited in the size of the training set, its results are fairly reasonable.

\textbf{LFE-Peaks} and \textbf{LFE-PPN} manage to outperform the state of the art in the low
recall/high precision zone of the PR graph. LFE-Peaks in particular is able to trade blows against LFE-PPN
and be competitive on its own against DROW3,
however it loses in mAP and AP @ 0.5\,m due to the length of its PR curve (stopping at less than 80\%
recall, while LFE-PPN is able to cross that threshold). These results clearly indicate that the 1D convolutional layers in LFE are capable of competently learning people-identifying features directly from range data. The PPN shows results that clearly validate the idea of replacing classical post-processing algorithms with a fully deep approach inspired by object detection,  however it still requires further tuning and improvements to be able to outperform the state of the art. Of note is its difference in performance depending on the associating distance, which indicates the need to further adjust people center proposal generation.

In terms of inference speed, we can observe large differences between processing times achieved by each detector. Some implementations (\texttt{leg\_detector} and our LFE-Peaks/LFE-PPN nodes) take less than 2ms to process each scan. Comparing these three detectors, there is a difference of about 0.3ms between LFE-PPN and the other two -- in other words, LFE-PPN is 15\% faster. DR-SPAAM takes around 14ms (similar to the time reported in \cite{Jia2020DRSPAAM}) regardless of temporal window size thanks to its auto-regressive architecture. DROW3's network architecture is slightly faster than DR-SPAAM's, taking around 13ms. Finally, PeTra takes around 28ms.
Some of these large differences could be attributed to a variety of reasons not necessarily related to model architecture, such as differences in the software technologies used. For instance, DROW3/DR-SPAAM's ROS node is fully implemented in Python and PyTorch (which in turn includes \texttt{rospy} serialization/deserialization overhead), while the rest are all implemented in C++. PeTra uses TensorFlow v1.x -- a long since deprecated branch that is no longer maintained, while our LFE/PPN nodes use ONNX Runtime -- a library specifically designed for fast inference times in deployed applications. In any case, this comparison involves real systems that are currently available for use by robotics researchers.

Finally, we qualitatively evaluate the detectors. We provide a video
\footnote{\url{https://youtu.be/czzddmL1pLI}}
showing the
laser scan sequence (with the reference image feed from the robot in the upper right corner),
and plotted circles corresponding to both the ground truth and each detector. We only plot
detections whose confidence is greater or equal than a given threshold matching the confidence
of the point at the Peak-F1 score of each detector.
Stills from the video can be seen in Fig.~\ref{fig:qualit}, showing several
types of environments: indoors (Scenes 1, 4, 6), outdoors (2, 3, 5), crowded
(4), with challenging geometry (1, 2, 6). The most interesting thing to note about
the detectors is their different ways of producing false positives. Certain kinds
of challenging geometry (such as pillars, fences or wall corners) can cause models
to incorrectly predict the presence of people. Models that do not aggregate information
from several scans tend to make more mistakes in these situations.

\section{Conclusions and future work}

We showcased a brand new dataset for people detection using 2D range finders called FROG, as
well as the process and tools we used to semi-automatically carry out the annotation process.
We also proposed our own deep learning based people detectors leveraging this data; and
afterwards we designed, implemented and carried out a benchmark intended to evaluate people
detectors using the FROG dataset. We obtained and reported results for a collection of
state-of-the-art detectors, and commented on the performance of each one. We can draw certain conclusions: 2D LiDAR-based person detection is still an open problem, and we hope we contribute to it through our dataset and our proposed models.

As future work, we intend to improve our models so that a fully
deep learning based approach (without non-deep pre-processing and classical post-processing steps)
can surpass the performance of existing models, besides already providing a faster implementation. Moreover, we plan to focus our efforts on the speed and usability of the models when executed directly on a real robot platform with low power on-device AI accelerators, as opposed to a separate desktop system with powerful hardware. Regarding the dataset, we recorded more sequences than we annotated. This opens the possibility of extending the dataset in the future, and using the new data as a hidden test set for competition purposes. Finally, we propose
further exploring the potential of self-supervised approaches \cite{Jia2021Person2DRange,sixthsense}, as well as fusing detection results from different sensor sources for a combined integral approach to people detection.


\bibliography{detector}

\end{document}